\DocumentMetadata{
      pdfversion=2.0,pdfstandard=ua-2,
      testphase={firstaid, math, title}
    }
\documentclass[sigconf]{acmart-tagged}
\AtBeginDocument{%
  }

\setcopyright{cc}
\setcctype{by}
\acmDOI{10.1145/3757279.3788669}
\acmYear{2026}
\copyrightyear{2026}
\acmISBN{979-8-4007-2128-1/2026/03}
\acmConference[HRI '26]{Proceedings of the 21st ACM/IEEE International Conference on Human-Robot Interaction}{March 16--19, 2026}{Edinburgh, Scotland, UK}
\acmBooktitle{Proceedings of the 21st ACM/IEEE International Conference on Human-Robot Interaction (HRI '26), March 16--19, 2026, Edinburgh, Scotland, UK}
\received{2025-09-30}
\received[accepted]{2025-12-23}

\newcommand{\para}[1]{\par\noindent \textbf{{#1}}}
\usepackage{amsmath}
\usepackage{tabularx}
\usepackage{multirow}
\usepackage{booktabs}
\usepackage{microtype}
\usepackage[ruled, noend]{algorithm2e} % For algorithms

\SetCommentSty{rmfamily}
\begin{document}

\title{Learning Contextually-Adaptive Rewards via Calibrated Features}

\author{Alexandra Forsey-Smerek}
\orcid{0000-0002-9324-2388}
\affiliation{%
  \institution{Massachusetts Institute of Technology}
  \city{Cambridge}
  \country{USA}
}
\email{aforsey@mit.edu}

\author{Julie Shah}
\orcid{0000-0003-1338-8107}
\affiliation{%
  \institution{Massachusetts Institute of Technology}
  \city{Cambridge}
  \country{USA}
}
\email{julie_a_shah@csail.mit.edu}

\author{Andreea Bobu}
\orcid{0000-0002-9507-7427}
\affiliation{%
  \institution{Massachusetts Institute of Technology}
  \city{Cambridge}
  \country{USA}
}
\email{abobu@mit.edu}

\begin{abstract}
 A key challenge in reward learning from human input is that desired agent behavior often changes based on context. For example, a robot must adapt to avoid a stove once it becomes hot. We observe that while high-level preferences (e.g., prioritizing safety over efficiency) often remain constant, context alters the \textit{saliency}--or importance--of reward features. For instance, stove heat changes the relevance of the robot’s proximity, not the underlying preference for safety. Moreover, these contextual effects recur across tasks, motivating the need for transferable representations to encode them. Existing multi-task and meta-learning methods simultaneously learn representations and task preferences, at best \textit{implicitly} capturing contextual effects and requiring substantial data to separate them from task-specific preferences. Instead, we propose \textit{explicitly} modeling and learning context-dependent feature saliency separately from context-invariant preferences. We introduce \textit{calibrated features}--modular representations that capture contextual effects on feature saliency--and present specialized paired comparison queries that isolate saliency from preference for efficient learning. Simulated experiments show our method improves sample efficiency, requiring 10x fewer preference queries than baselines to achieve equivalent reward accuracy, with up to 15\% better performance in low-data regimes (5–10 queries). An in-person user study (N=12) demonstrates that participants can effectively teach their personal contextual preferences with our method, enabling adaptable and personalized reward learning.
\end{abstract}
\begin{teaserfigure}
\vspace{-0.5em}\centering\includegraphics[width=\textwidth, alt={Conceptual depiction of framework. Users first respond to contextual feature queries to teach calibrated features then train rewards on top of these calibrated features. Figure illustrates how the same three trajectories might be valued differently when contexts or user preferences change.}]{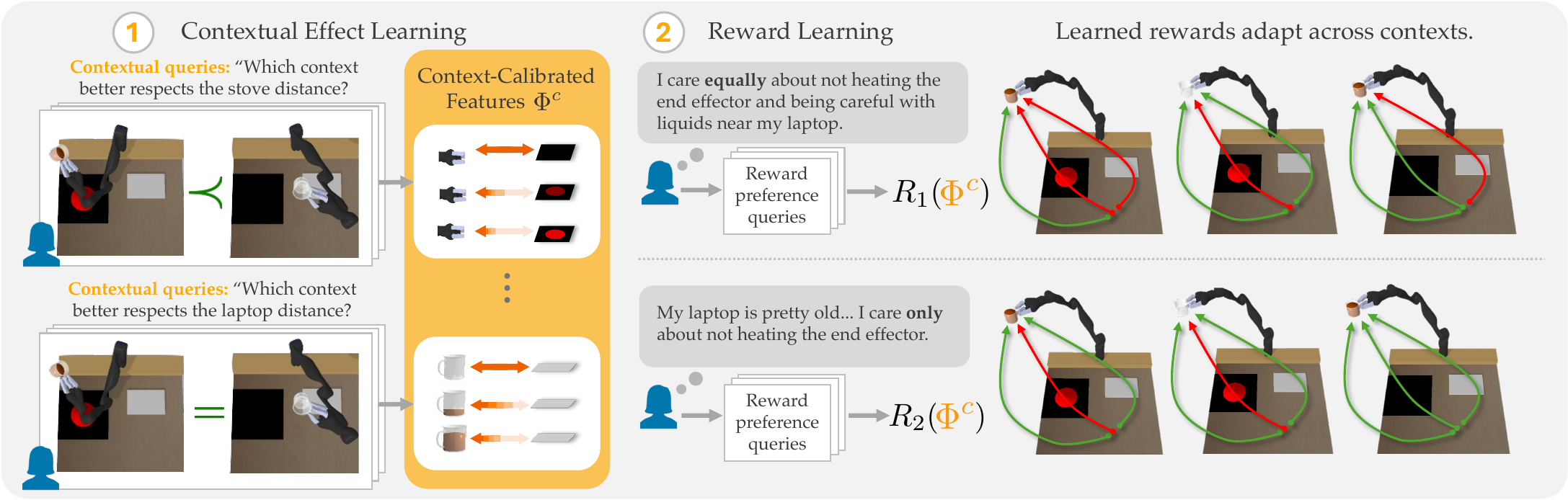}
    \vspace{-2em}
    \caption{
    We learn representations that capture how context affects reward feature saliency. (Step 1) Contextual queries  reveal when context changes feature importance: 
     stove distance is better respected when the stove is off (top), whereas laptop distance is equally respected whether the cup is empty near the laptop or full but far from it (bottom).
    (Step 2) Calibrated features enable learning contextually adaptive rewards (high rewards in green and low rewards in red).}
    \label{fig:conceptual}
    \Description{Conceptual depiction of framework. Users first respond to contextual feature queries to teach calibrated features then train rewards on top of these calibrated features. Figure illustrates how the same three trajectories might be valued differently when contexts or user preferences change.}
\end{teaserfigure}
\maketitle
\section{Introduction}
Imagine a robot helping you in the kitchen. When the stove is off, the robot can pass freely above it to reach the counter. 
As the stove heats up, the robot should keep its end effector at a safe distance from the heat, and when the stove is on high, that distance should be even greater.
This simple example highlights a core challenge for robot learning: \textit{contexts change, demanding different robot behaviors.}
Traditional reward learning frameworks would treat these scenarios as separate tasks with distinct rewards, even though they stem from the same underlying preference for safety. 
We argue that what actually changes across contexts is not the preference itself (e.g., safety over efficiency), but the relative importance--or \textit{saliency}--of reward features.
In our example, the stove's temperature doesn't alter the human's desire for safety, but rather modulates how much the robot's stove distance \textit{matters} in achieving it. Critically, these contextual effects on feature saliency are often consistent across rewards and independent of specific preferences: whether a user prioritizes efficiency or safety, a cold stove renders stove distance irrelevant. 
This view aligns with cognitive science theories of hierarchical value construction~\cite{o2021hierarchical}, and motivates the need for reusable representations that capture contextual effects on feature saliency, independent of task-specific rewards. 
Such representations would enable efficient learning of rewards that adapt across contexts. 

Predefined feature representations rarely capture contextual effects and are difficult to personalize, motivating approaches that learn representations from data. 
Multi-task and meta inverse reinforcement learning (IRL) methods aim to learn transferable structure by training shared representations across multiple tasks~\cite{yamada2022task, yu2019meta}. In principle, they could capture shared contextual effects--for example, learning that stove temperature consistently influences distance preferences across users' reward functions. In practice, however, they must learn two things simultaneously: how context modulates feature saliency and how users weight different features. As a result, contextual effects are captured only \textit{implicitly}, entangled with task-specific trade-offs, and disentangling the two requires large amounts of training data.
Our key insight is that robots should instead \textit{explicitly} model and learn how context affects feature saliency via targeted human feedback, rather than hoping these patterns will emerge with ever-growing data.

We propose learning these relationships through \textit{calibrated features}--representations that incorporate contextual effects on task reward features. Calibrated features explicitly capture how contextual elements (e.g., stove temperature) should affect task feature values. For example, when the stove is off, the learned calibrated feature should map any distance from the stove to the same value--reflecting that being 2 or 10 steps away makes no practical difference when there's no heat to avoid. We learn calibrated features as a function of the base feature and the input state. Therefore, the representations capture how the inherent state context affects feature saliency. By capturing contextual effects individually for each feature, we learn contextually-aware representations that can be modularly reused and composed in new rewards.

We introduce a novel type of human input for learning calibrated features, \textit{contextual feature queries}. These paired comparison queries target how context affects feature saliency. Each query presents a user with two robot states, and asks them to select the context in which the robot better respects a particular feature. In Fig.~\ref{fig:conceptual}, for the ``stove distance'' feature, the user selects the second state with the stove off over the first state where the robot is dangerously close to the hot stove. For the ``laptop distance'' feature, the user labels the states equivalent since in one state the cup is empty and the laptop safe regardless of the robot’s position, and in the other state the cup contains liquid but is held far from the laptop. By focusing human feedback directly on these contextual relationships, we can efficiently learn how context reshapes task features.

In summary, our contributions are three-fold. First, we introduce a framework for learning representations that explicitly model contextual effects through \textit{calibrated features}. 
Second, we propose a method for learning calibrated features using targeted \textit{contextual feature queries}. Finally, we empirically show our approach improves sample efficiency in three complex reward learning tasks in robotic manipulation, outperforming multi-task baselines. A user study validates that real people can respond to contextual feature queries, and reveals diverse user contextual preferences, emphasizing the importance of sample efficiency for learning from individual input. 
\vspace{-1em}
\section{Background}
\para{Learning from Human Input.} Recent advances in robot learning from human input primarily revolve around two paradigms: Inverse Reinforcement Learning (IRL) \cite{ng2000algorithms} and Behavior Cloning (BC). 
BC learns a state-conditioned policy from expert actions, making it vulnerable to distribution shift, whereas IRL infers a reward function that explains observed behavior and generalizes to new states.
Various forms of human input for IRL have been explored, including demonstrations \cite{abbeel2004apprenticeship, ziebart2008maximum}, preferences \cite{christiano2017deep, sadigh2017active}, and corrections \cite{bajcsy2017learning, jain2015learning}. Beyond reward inference, prior work has also investigated representation learning from human input. Existing methods aim to learn transferrable reward representations, either implicitly learning encodings  \cite{bobu2023sirl} or 
explicitly learning features \cite{bobu2022inducing}, though both without modeling contextual effects. Other recent work uses human input to learn safety constraints as control barrier functions to specify unsafe robot behaviors (e.g., cup tilt angle limit) \cite{yang2024enhancing}. Our work is complementary; instead of enforcing hard safety constraints, we provide granular control by attenuating feature values.  

\para{Modeling Context.} Limited prior work explicitly models contextual effects on learned behaviors.  
Related work \cite{ghosal2023contextual} models contextual effects as a binary switch determining which features are task-relevant. Rather than switching features on and off, we learn representations that capture the effect of context on saliency in a continuous, gradual way. 
Other work \cite{nishi2020fine, sodhani2021multi} structures multi-task rewards as a function of both features and context, assuming explicit access to task context either as metadata \cite{sodhani2021multi} or discrete values \cite{nishi2020fine}. In contrast, we don't assume access to the context, instead learning which state elements compose the relevant context. 

\para{Meta and Multi-Task Reward Learning.}
Multi-task and meta-IRL approaches aim to learn representations that transfer across tasks.
Early multi-task IRL methods focused on clustering heterogeneous demonstrations and learning distinct rewards \cite{babes2011apprenticeship, dimitrakakis2012bayesian, choi2012nonparametric}, limiting information sharing across tasks. Following work explored more complex reward representations with methods for learning shared task representations \cite{nishi2020fine, yamada2022task}. Developments in meta-IRL began with applications of MAML \cite{finn2017model} to tabular settings \cite{xu2019learning} and known task distributions \cite{gleave2018multi}, followed by the introduction of latent task variables for policy conditioning \cite{yu2019meta, seyed2019smile}. Recent work has addressed long-horizon planning \cite{chen2023multi} and complex environments where maintaining a distribution over tasks is intractable \cite{wang2021meta}. Given prior work showing multi-task pre-training can match or outperform meta-RL while being more computationally efficient \cite{zhao2022effectiveness}, we compare our method to a multi-task baseline \cite{yamada2022task}. Finally, multi-objective RL (MORL) considers a setting like ours, modeling rewards as functions of base objectives, but typically focuses on Pareto-optimal policies \cite{van2014multi}. State-of-the-art methods for learning MORL objectives \cite{mu2025preference} assume known training rewards, which is more restrictive than our baseline. Unlike all of these approaches, our method doesn't require multiple training tasks.
\section{Approach}
\subsection{Problem formulation}
We are interested in learning representations that adapt to context for better downstream reward learning. % when executing a task. 
We model our problem as a Markov decision process
$(\mathcal{S}, \mathcal{A}, \mathcal{T}, R)$, with states 
$s \in \mathcal{S} \subseteq \mathbb{R}^d$, actions $a \in \mathcal{A}$, transition $\mathcal{T}: \mathcal{S} \times \mathcal{A} \times \mathcal{S} \rightarrow [0, 1]$, and rewards $R:\mathcal{S} \times \mathcal{A} \rightarrow \mathbb{R}$. To express the reward function, we assume the agent has $K$ base features $\phi(s): \mathbb{R}^d \rightarrow [0,1]$\footnote{We normalize base and calibrated features to $[0,1]$ for training stability, but in general they may lie in $\mathbb{R}^+$.}, which represent useful task aspects (e.g., laptop distance) and can be hand-crafted~\cite{ziebart2008maximum}, learned \cite{bobu2022inducing}, or even LLM-generated~\cite{peng2024adaptive}. We denote the feature set as $\Phi(s) = \{\phi_i(s)\}_{i=1}^K$, where $\phi_i(s)$ is the $i$-th feature.

We assume context is already captured in the state as elements $s_c \in s$ that affect the user's desired behavior;
for instance, stove temperature affects desired stove distance and cup fullness affects desired cup orientation. 
Since in general it's unrealistic to assume the robot knows which state elements constitute context or how they affect behavior, we learn representations that infer these relationships directly from the state $s$.
We formulate this problem as learning a representation $f_{\psi}(\Phi(s),s) : [0,1]^K \times \mathbb{R}^d \rightarrow \mathbb{R}^h$, parameterized by $\psi$. For brevity, we denote this mapping as $f_{\psi}(s)$. The role of $f_{\psi}$ is two-fold; it must capture which state elements comprise the relevant context $s_c$ and how these contextual elements should influence the reward. 
We can then learn a reward function on top of the learned representation, $R_{\theta}(f_{\psi}(s)): \mathbb{R}^h \rightarrow \mathbb{R}$, parameterized by $\theta$, producing a reward that can adapt across contexts.

One way to learn these representations is through multi-task or meta-learning: train $N$ different reward functions $\{R_{\theta_i}\}_{i=1}^N$ that share a similar underlying representation $f_\psi(s)$. In this setup, the model must jointly learn task-specific parameters $\{\theta_i\}_{i=1}^N$ and shared representation parameters $\psi$. Disentangling shared contextual structure from task-specific preferences requires many diverse training tasks, making these approaches data-hungry and inefficient.
Moreover, these methods learn representations that can't be selectively reused or combined with new features in new rewards without retraining. We identify that end-user input can be more effective when directed toward explicitly describing contextual effects, rather than providing only general preference signals. Furthermore, if we learn feature-specific representations, these can be modularly reused and composed in downstream rewards.
\vspace{-0.5em}
\subsection{Structuring reward with calibrated features}
We propose modeling the effects of context using an intermediate representation on top of each contextually affected feature. We present \textit{calibrated features} as mappings from a base feature and state to a calibrated feature value, $\phi_{\psi_i}^c(\phi_i(s),s): [0,1] \times \mathbb{R}^d \rightarrow [0,1]$, where the $i$-th calibrated feature, a function approximator parameterized by $\psi_i$, corresponds with the $i$-th base feature. We denote $\phi_{\psi_i}^c(\phi_i(s), s)$ as $\phi_{\psi_i}^c(s)$ for notational convenience. The $i$-th calibrated feature serves to capture which elements of the state should affect the saliency of the $i$-th base feature, and how. 
A learned calibrated feature would take as input the state and a base feature, such as the cup angle, and infer which state elements should condition the cup angle feature value (cup fullness) and how (saliency of cup angle is positively correlated with cup fullness). Importantly, we do not learn a function that outputs feature saliency, but one that conditions an input feature on the state context, producing a new, reshaped feature.  Because each calibrated feature depends only on its base feature and the state, it can be reused compositionally with new features as long as the state input remains the same.

Since our approach operates on semantically meaningful features, we assume that users can indicate which features should be contextually affected. This assumption is further discussed in Sec.~\ref{sec:discussion}.
For a given task, we compose the representation $\Phi^c(s) = \{\phi_i^c(s)\}_{i=1}^K$ as a mix of calibrated and uncalibrated features, swapping in calibrated features when contextual effects exist,
\begin{equation}
\phi_i^c(s) =
\begin{cases}
\phi_{\psi_i}^c(\phi_i(s), s), & \text{if $c_i$}, \\[6pt]
\phi_i(s), & \text{otherwise}, 
\end{cases}
\end{equation}
where $c_i\in\{0,1\}$ is a binary indicator provided by the end user denoting if the $i$-th feature is contextually affected.
We learn rewards on top of this representation, $R_{\theta}(\Phi^c(s)): [0,1]^K \rightarrow \mathbb{R}$. When the reward is linear, parameters $\theta$ maintain their traditional role quantifying the trade-off \textit{between} features, while calibrated features capture contextual variation \textit{within} a feature. Shown in Fig.~\ref{fig:conceptual}, reward parameters describe the trade-off between staying away from hot objects and being careful with liquids near technology, while calibrated features reshape how each feature influences the corresponding higher-level concept, given the context.
\vspace{-0.5em}
\subsection{Learning calibrated features}
We leverage paired comparison queries for learning calibrated features, drawing inspiration from prior work on learning features from human input \cite{bobu2022inducing}. Unlike traditional comparison approaches that ask users to select the highest-reward state, we solicit input on how changes in state influence the saliency of an individual feature. We present users with \textit{contextual feature queries}, where they are shown two states and asked to pick in which state a specific feature is better respected, focusing their attention on that feature alone. 

For each contextually affected feature $\phi_i$, we collect a set of $M_i$ paired comparisons $\mathcal{D}_{\phi_i} = \{(s_1^{m}, s_2^{m}, y^{m})\}_{m=1}^{M_i}$. Paired comparison labels $y \in \{0, 0.5, 1\}$ let the user indicate whether the feature is more respected in $s_1 \ (y=1)$, in $s_2 \ (y=0)$, or equivalently across states $(y=0.5)$. Equivalence information is crucial for identifying which portions of the state should not affect feature saliency, and contexts in which feature saliency should remain static despite changes in input state value. We split the dataset $\mathcal{D}_{\phi_i}$ into two subsets which contain all queries labeled with equivalence, $\mathcal{D}_{\phi_i}^{\textnormal{equiv}} = \{ (s_1, s_2, y) \in \mathcal{D}_{\phi_i}\mid y=0.5 \}$
and all others, $\mathcal{D}_{\phi_i}^{\textnormal{pref}} = \{ (s_1, s_2, y) \in \mathcal{D}_{\phi_i}\mid y \neq 0.5 \}$. 

Following \cite{bobu2022inducing} we model human feature comparisons using the Bradley-Terry framework \cite{bradley1952rank}, 
\begin{equation} \label{eq:BT}
    P_{\psi_i}(s_1 \succ s_2) =  
     \frac{\exp(\phi'_{\psi_i}(s_1))}{\exp(\phi'_{\psi_i}(s_1)) + \exp(\phi'_{\psi_i}(s_2))},
\end{equation}
and use a cross-entropy loss over a dataset $\mathcal{D}$ of paired comparisons, 
\begin{equation} \label{eq:cross-ent}
\ell(\psi_i; \mathcal{D}) = \sum_{(s_1, s_2, y) \in \mathcal{D}} \left[ -y \log P_{\psi_i}(s_1 \succ s_2) - (1 - y) \log P_{\psi_i}(s_1 \prec s_2) \right].
\end{equation}
The final loss equation becomes,
\begin{equation} \label{eq:final-loss}
\mathcal{L}(\psi_i) =  
\frac{1}{\lvert\mathcal{D}_{\phi_i}\rvert} \left(\lambda_{\text{FE}}\cdot \ell(\psi_i; \mathcal{D}_{\phi_i}^{\text{equiv}}) + \ell(\psi_i; \mathcal{D}_{\phi_i}^{\text{pref}})\right),
\end{equation} 
where $\lambda_{\text{FE}}$ is a hyperparameter that scales the contribution of equivalence-labeled pairs. 
Alg.~\ref{pseudo-code} details our learning method.
\subsection{Learning rewards with calibrated features}
In this work, we learn reward from preferences, but our framework could be adapted to any other reward learning approach, such as demonstrations~\cite{finn2016gcl} or corrections~\cite{bajcsy2017learning}.
We ask the human reward preference queries, presenting them with two states and asking them to select which state they prefer. We collect a set of paired comparisons $\mathcal{D}_{\theta} = \{(s_1^{m}, s_2^{m}, y^{m})\}_{m=1}^{M_{\text{REW}}}$, where $M_{\text{REW}}$ is the number of reward comparisons. We enable equivalence labels such that $y \in \{0, 0.5, 1\}$, indicates if $s_1 $ is preferred $(y=1)$, if  $s_2 $ is preferred $(y=0)$ or if the states are viewed as equivalent $(y=0.5)$. Again, paired comparisons are split into subsets which contain all queries labeled with equivalence, $\mathcal{D}_{\theta}^{\text{equiv}} = \{ (s_1, s_2, y) \in \mathcal{D}_{\theta}\mid y=0.5 \}$
and all others, $\mathcal{D}_{\theta}^{\text{pref}} = \{ (s_1, s_2, y) \in \mathcal{D}_{\theta}\mid y \neq 0.5 \}$. We model human reward preferences with the same Bradley-Terry framework (Eq.~\ref{eq:BT}) and employ cross entropy loss (Eq.~\ref{eq:cross-ent}), resulting in the final loss, 
\begin{equation} \label{eq:final-loss-rew}
\mathcal{L}(\theta) =  
\frac{1}{\lvert \mathcal{D}_{\theta}\rvert} \left(\lambda_{\text{RE}}\cdot \ell(\theta; \mathcal{D}_{\theta}^{\text{equiv}}) + \ell(\theta; \mathcal{D}_{\theta}^{\text{pref}})\right),
\end{equation} 
where $\lambda_{\text{RE}}$ scales the contribution of equivalence-labeled pairs.
\begin{figure*}[!t]    \centering\includegraphics[width=0.95\textwidth, alt={Left panel shows three manipulation environments. The robot holds a separate object in each (block, utensil, or a cup). Right panel shows calibrated feature evaluated at all states in a point cloud above a table top environment. Calibrated features shown for block, stove, utensil, and cup features across a range of the relevant contextual element.}]{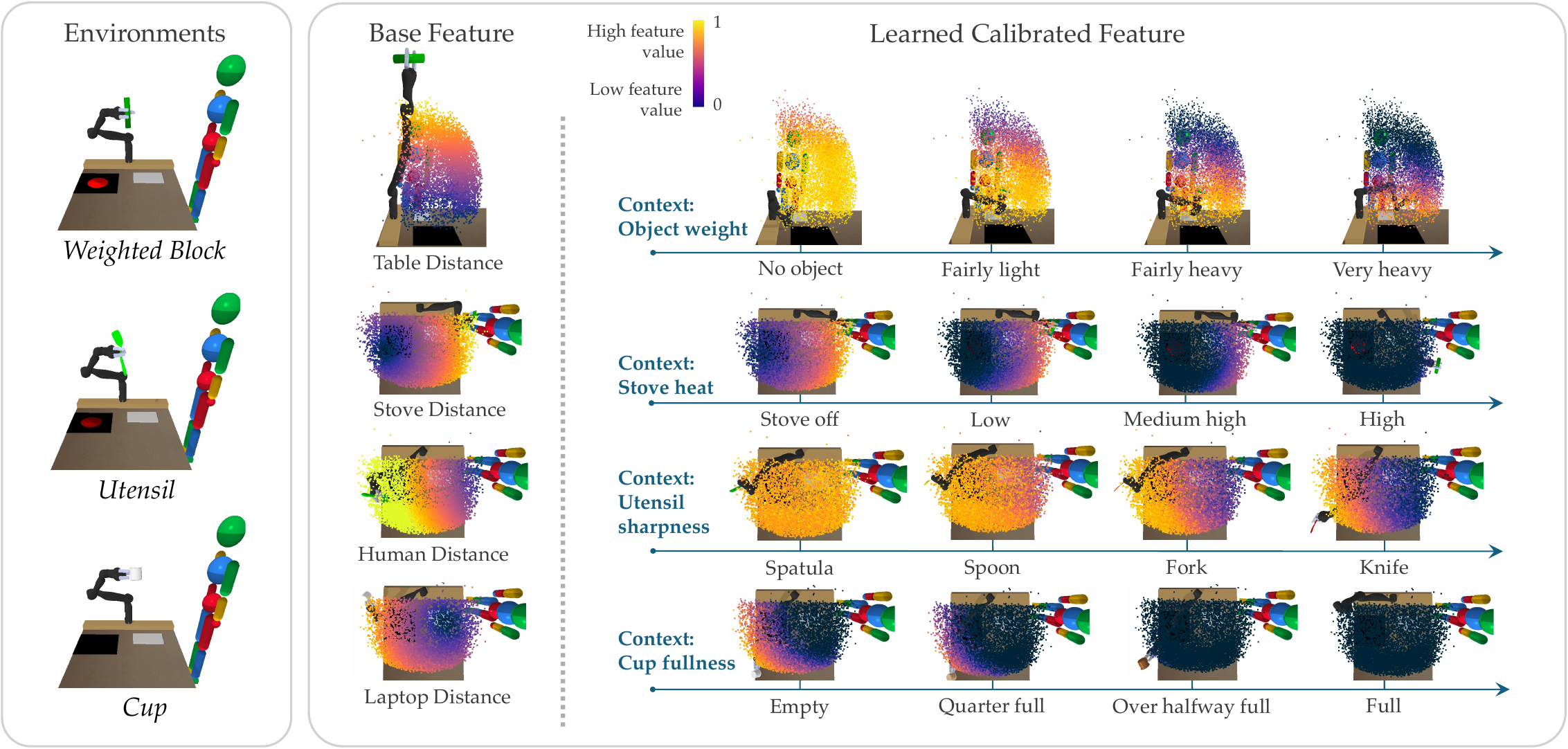}
    \caption{Left: three manipulation environments. Right: calibrated features learned from simulated human input. Learned contextual effects reshape the base feature as the relevant contextual element changes. Point clouds show 8k EE locations, normalized over the full range of the relevant contextual element. Irrelevant contextual elements fixed to zero. 
    }.
    \Description{Left panel shows three manipulation environments. The robot holds a separate object in each (block, utensil, or a cup). Right panel shows calibrated feature evaluated at all states in a point cloud above a table top environment. Calibrated features shown for block, stove, utensil, and cup features across a range of the relevant contextual element.}
    \label{fig:calib_feats}
    \vspace{-1em}
\end{figure*}
\vspace{-0.5em}
\section{Experiments in Simulation}
We first evaluate our method using simulated human input to objectively measure reward accuracy against ground truth. Our experiments each consist of two stages: (1) representation learning, and (2) reward learning using learned representations. We compare rewards learned using calibrated features versus multi-task baseline representations, investigating whether: (1) explicitly capturing contextual effects using calibrated features yields more accurate rewards than baselines, and (2) learned calibrated features can be effectively composed to learn new downstream rewards.
\vspace{-0.5em}
\subsection{Environments} We designed three robotic manipulation environments simulated with PyBullet \cite{coumans2016pybullet} featuring a Jaco robot in a tabletop setting (Fig.~\ref{fig:calib_feats}). Environments were constructed to reflect everyday scenarios where desired behavior may adapt based on varying environmental conditions and object properties, such as the desire to carry heavier objects closer to the table and farther from electronics, or keep sharper objects farther from a human and pointed away. Preferred adaptations can be personal, reflecting individual differences in traits like caution or comfort.
Each environment contains three task-relevant features, which enable rich reward composition further complexified by contextual effects.
The continuous state for each environment is 23-dimensional: 3D end-effector (EE) position and rotation matrix (9 values), 3D positions of three fixed objects (laptop, stove, and human), and two environment-specific contextual elements. 
\begin{algorithm}[!t]
\DontPrintSemicolon
  \caption{Learning Calibrated Features}\label{pseudo-code}
  \KwIn{Features $\Phi$, $\{M_i\}_{i=1}^K$ number of queries per feature}  
  \For{$i \gets 1$ \KwTo $\lvert\Phi\rvert$}{
        $c_i\gets$  QueryHuman \tcp{ask if $\phi_i$ contextually affected}
        \If{$c_i$}{
             $\mathcal{D}_{\phi_i} \gets \{\}$ \;
             \For{$j \gets 1$ \KwTo $M_i$}{
                % \tcc{sample pairs; we use uniform but other strategies can be used}
                $(s_1, s_2) \gets \textsc{Uniform} (\mathcal{S})$ \tcp{sample (any strategy)}
                $y \gets$ ContextualQuery($\phi_i, s_1, s_2$) \tcp{ask human}
                $\mathcal{D}_{\phi_i} \gets \mathcal{D}_{\phi_i} \cup (s_1, s_2, y)$\;
             }
             $\phi^c_{\psi_i}(s) \gets \text{train\_calib\_feat}(\mathcal{D}_{\phi_i})$ \tcp{via Eq.~\ref{eq:final-loss}}
             % $\vec{\phi}' \gets \vec{\phi}' \cup \phi'_{\psi_i}$\;
             $\phi^c_i(s) \gets \phi^c_{\psi_i}(s)$
        }
        \Else{
            $\phi_i^c(s) \gets \phi_i(s)$
        }
  }
\end{algorithm}
For evaluation and simulating human responses, we designate these two state elements as the ground truth contextual factors that influence feature saliency (but don't assume knowledge of the subset at training). 
The three environments are:\\
\textit{Weighted Block}. The robot carries a block with variable weight. Features are: (1) \textit{stove}, EE xy-planar stove distance, (2) \textit{laptop}, EE xy-planar laptop distance, and (3) \textit{table}, EE height above table. Contextual elements are stove heat and block weight. \\ 
\textit{Cup}. The robot carries a cup with variable fullness. Features are: (1) \textit{stove}, (2) \textit{laptop}, and (3) \textit{cup angle}, defined as the projection of the EE's x-axis (cup’s upright axis) onto the world vertical axis. Contextual elements are stove heat and cup fullness.\\
\textit{Utensil}. The robot carries a utensil with variable sharpness. Features are: (1) \textit{stove}, (2) \textit{human}, EE xy-planar human distance, and (3) \textit{point}, defined as the projection of the end-effector’s x-axis (utensil’s pointing direction) onto the unit vector pointing from the EE to the human. Contextual elements are stove heat and utensil sharpness.\\
For more details on base feature implementation, see App.~\ref{app:base-feat-def}.
\para{Human Simulation.}
We simulate human responses by defining ground truth calibrated feature functions and reward models. We define seven ground truth calibrated feature functions (see details in App.~\ref{app:gt-calib-feat-def}), which serve as examples of desired contextual relationships (context$\to$feature): (1) stove heat$\to$\textit{stove}, (2) block weight$\to$\textit{laptop}, (3) block weight$\to$\textit{table}, (4) cup fullness$\to$\textit{laptop}, (5) cup fullness$\to$\textit{cup angle}, (6) utensil sharpness$\to$\textit{human}, and utensil sharpness$\to$\textit{point}. 
Ground truth rewards are linear combinations of calibrated features, representing tradeoffs such as it being more important to keep sharp objects away from a person than to avoid heating up the EE. 
We simulate human responses to state comparison queries using the Bradley-Terry model of human decision making~\cite{bradley1952rank}.
We consider states equivalent if their calculated calibrated feature values or rewards fall within an empirically determined threshold (see App.~\ref{app:sim_human-threshold} for threshold selection).
\vspace{-0.5em}
\subsection{Experimental Design} 
\begin{figure*}[!ht]
    \centering
\includegraphics[width=0.95\textwidth, alt={Line graph showing reward accuracy against number of reward preference queries. Our method shows higher performance at all non-zero query counts, outperforming all baseline methods. The gap between our method and the best baseline is most pronounced in low data regimes (5-10 preference queries) and closes toward the higher query counts (50 queries).}]{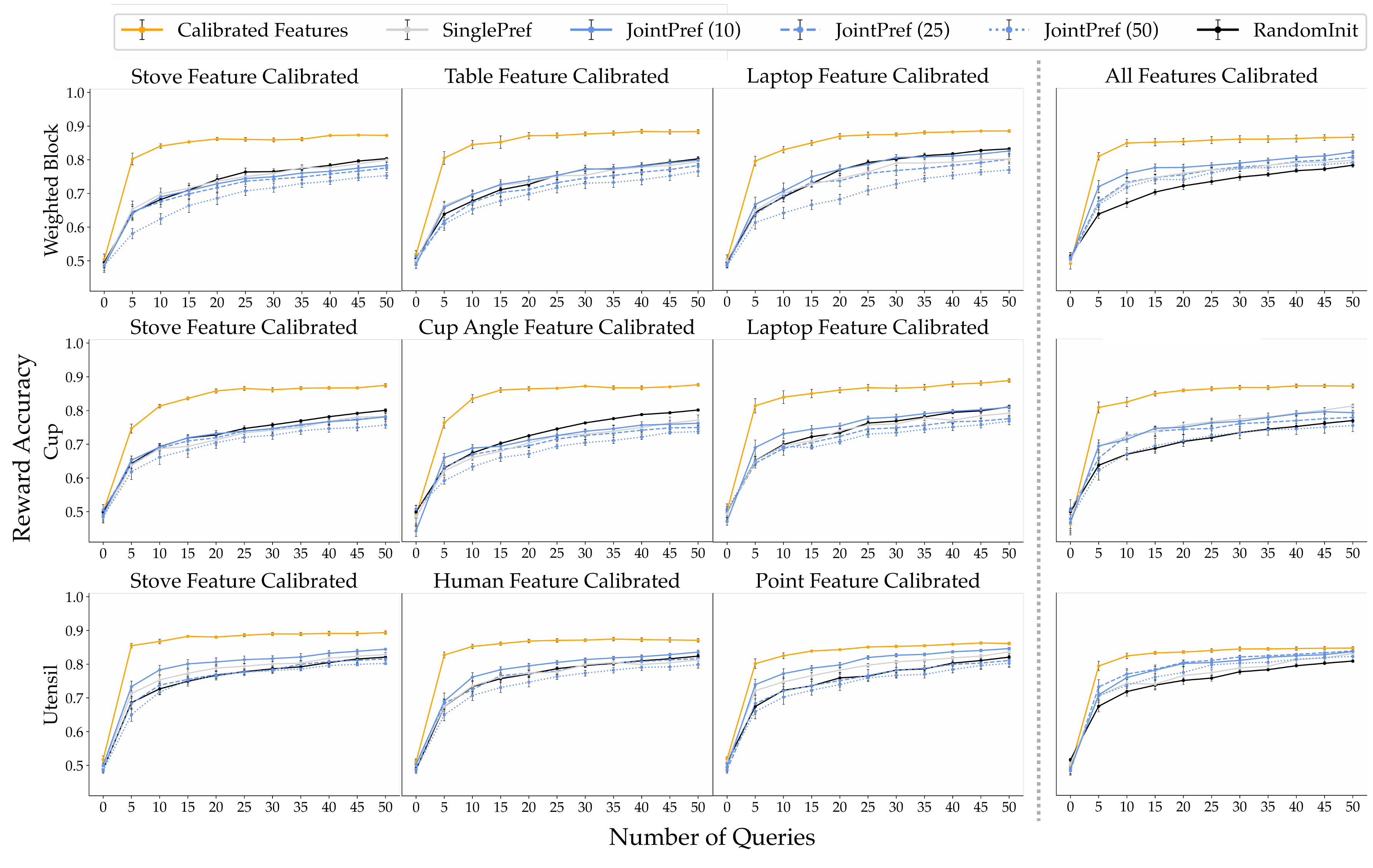}
    \caption{Exp. 1 \& 2 reward accuracy results. Results shown for each environment (rows) when learning rewards that capture contextual effects on one of the three environment features (columns 1-3) or all three environment features (column 4).}
    \label{fig:single_calib}
    \Description{Line graph showing reward accuracy against number of reward preference queries. Our method shows higher performance at all non-zero query counts, outperforming all baseline methods. The gap between our method and the best baseline is most pronounced in low data regimes (5-10 preference queries) and closes toward the higher query counts (50 queries).}
\end{figure*}
\begin{figure*}[h!]
    \centering    \includegraphics[width=0.95\textwidth, alt={Bar graph showing induced trajectory reward results for each environment for each single calibrated feature and all calibrated features. Our method outperforms all baseline methods across all experiments.}]{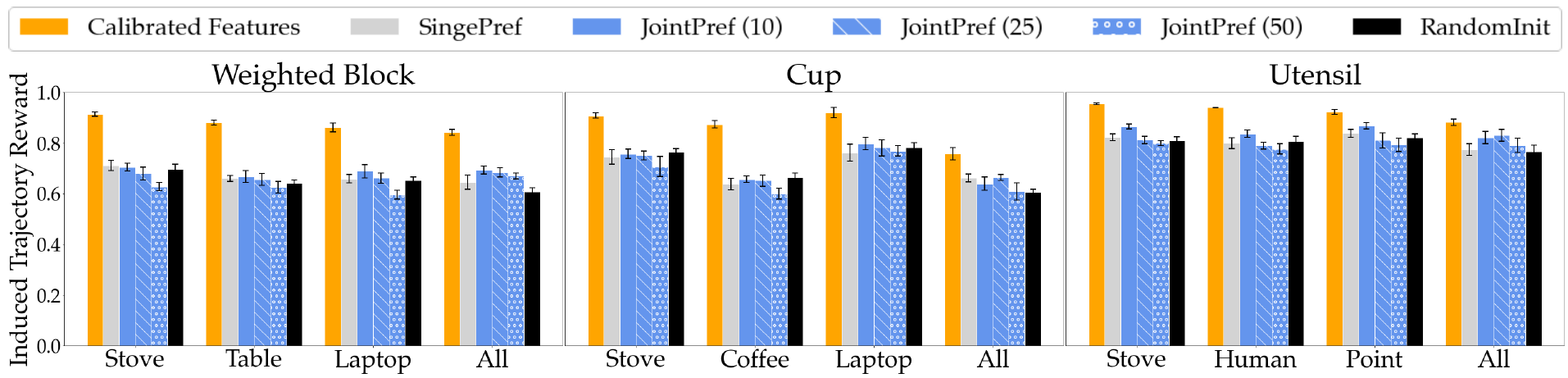}
    \caption{Exp 1 \& 2 induced trajectory reward results,  showing ground truth reward of trajectories selected by learned rewards. Results shown for each environment for rewards in which a single or all three environment features are contextually-affected.}
    \label{fig:induced-traj}
    \Description{Bar graph showing induced trajectory reward results for each environment for each single calibrated feature and all calibrated features. Our method outperforms all baseline methods across all experiments.}
    \vspace{-0.5em}
\end{figure*}
\para{Manipulated Variables.} We compare four representation learning methods when either a single feature or all features are contextually affected. Each method is allotted a budget of $M_{\text{REP}}$ queries. See App.~\ref{app:imp-details} for implementation and training details. \\
(1) \textbf{Calibrated Features}. We assume access to the set of contextually-affected features and train a calibrated feature for each, evenly dividing $M_{\text{REP}}$ contextual feature queries across them. 
\\  
(2) \textbf{JointPref}. Following prior work showing multi-task pre-training for representation learning to be equally or more effective than meta-RL \cite{zhao2022effectiveness}, we compare against \textit{JointPref}, a multi-task baseline \cite{yamada2022task} where representations are learned by pre-training $N$ reward heads with shared initial layers.
We adapt their approach to use $M_{\text{REP}}/N$ preference queries to train each reward head, testing implementations with $N=\{10,25,50\}$ (\textit{JointPref}-\textit{N}). Model input is the vector of base features concatenated with the state. Shared layers compress the input to a 7-dimension embedding.\\
(3) \textbf{SinglePref}. We implement \textit{JointPref} with $N=1$, where the sole training preference uniformly weights features.\\
(4) \textbf{RandomInit}. We include a randomly initialized representation with the same architecture as the multi-task baseline shared layers. This offers a lower bound for the benefits of pre-training.
\begin{figure*}[!t]
    \centering    \includegraphics[width=0.9\textwidth, alt={Box and whisker plots showing user Likert Scale responses and Ranking responses. A separate box is shown for each feature for each number of training queries (0, 50, 25, and 100). Likert scale scores increase as the number of training queries increases, while rankings decrease with increased training queries (given lower is better).}]{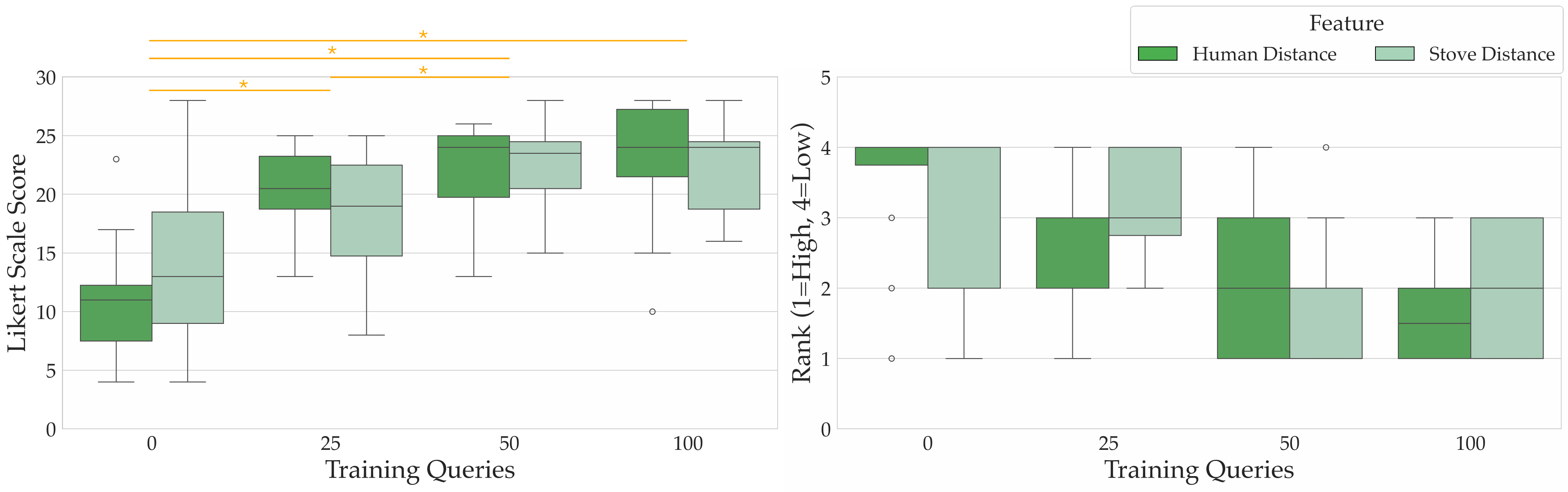}
    \caption{Subjective user ratings and model ranking improved with increasing training queries (*p<0.05). The \textit{human} feature benefits from 100 vs. 50 queries, while the \textit{stove} feature plateaus after 50 queries, highlighting that query requirements vary with feature complexity. Boxes show 25th–75th percentiles and median, whiskers mark 1.5$\times$IQR, and points mark outliers.}
    \label{fig:hpe_data}
    \Description{Box and whisker plots showing user Likert Scale responses and Ranking responses. A separate box is shown for each feature for each number of training queries (0, 50, 25, and 100). Likert scale scores increase as the number of training queries increases, while rankings decrease with increased training queries (given lower is better).}
    \vspace{-0.75em}
\end{figure*}
\para{Metrics.} We evaluate the quality of the learned representations by learning 10 held-out test rewards on top of each representation type. We test both freezing and fine-tuning representations during reward learning, reporting the stronger variant for each method (our method with freezing, baselines with fine-tuning; see App.~\ref{app:frozen-results}). We compare the accuracy of rewards learned with varying amounts of randomly sampled preference queries per test reward, $M_{\text{REW}}$. To ensure evaluation on separate tests states from those the robot is trained on, we generate 10k possible states per environment which we split into an 80/20 train/test distribution (repeated over 6 seeds). To generate training queries, we randomly sample $M_{\text{REP}}$ representation learning queries and $M_{\text{REW}}$ reward learning queries from the distribution of 8k possible train states. Reward accuracy is then measured as the average win rate across all test rewards evaluated over 1k state pairs randomly sampled from the test states. We additionally evaluate the quality of trajectories induced by rewards learned with a fixed low-data preference query budget ($M_{\text{REW}}=10$). Across 100 generated trajectory sets (10 contexts × 10 start–goal pairs), each containing 100 trajectories, we compute the ground-truth reward (normalized within each set) of the  trajectory ranked the highest by the learned reward. We report this value averaged over all trajectory sets and all test rewards.

\para{Exp. 1: Single contextually affected feature.} We first test whether explicitly learning contextual effects with calibrated features provides an advantage over implicit capture by baselines. We examine cases where only one of the three reward features is affected by context. 
Each method is trained on $M_{\text{REP}}=100$ pre-training queries, chosen based on model convergence (see App.~\ref{app:rep-learning} for an in-depth analysis of convergence properties of our method and baselines).
\para{Exp. 2: Multiple contextually affected features.} We next investigate if learned calibrated features can be composed to learn accurate downstream rewards using scenarios where all three reward features vary with context. We accordingly scale the number of pre-training queries to $M_{\text{REP}}=300$. 
\vspace{-0.5em}
\subsection{Results}
\para{Qualitative analysis of learned calibrated features.}
Fig.~\ref{fig:calib_feats} compares base features to learned calibrated features across changes in the relevant contextual element. Results show that learned representations successfully capture meaningful feature shaping functions that appropriately calibrate features based on context. In the first row, the representation adapts the base \textit{table} feature to capture the effect of object weight, i.e., the heavier the object, the closer to the table it should be carried. The reshaped feature values encode the concept of safe carrying--values are almost uniform when no object is carried, but increasing object weight produces a sharper contrast in values at higher and lower positions.  
A similar trend appears across all rows: increasing the stove heat, utensil sharpness, and cup fullness affect the \textit{stove}, \textit{human}, and \textit{laptop} features, respectively.

\para{Exp. 1: Single calibrated feature.}
Fig.~\ref{fig:single_calib} columns 1-3 show learned reward accuracy when a single feature is contextually affected. Our method consistently outperforms the best-performing baseline across all experiments, with up to 15\% improvement in low-data regimes (5-10 pref queries). Fig.~\ref{fig:induced-traj} similarly shows trajectories induced by rewards learned using our method outperform the best-performing baseline when a single feature is calibrated. The best-performing baselines across metrics are often \textit{JointPref} (10) and \textit{SinglePref} models, although their performance is generally similar to random representation initialization. 
This indicates that in low-data regimes, implicit representation learning struggles to capture useful structure. This is unsurprising, given \textit{SinglePref} lacks the signal required to distinguish task-specific reward from general structure. Meanwhile, multi-task variants spread 100 representation queries across all rewards (10 queries per reward for \textit{JointPref} (10), 4 for \textit{JointPref} (25) and 2 for \textit{JointPref} (50)), leaving too little data to distinguish shared contextual effects from task-specific rewards. 

\para{Exp. 2: Calibrated feature composition.} 
Results (Fig.~\ref{fig:single_calib} \& ~\ref{fig:induced-traj}) demonstrate that our approach continues to outperform the best baseline across metrics, though with a narrower performance gap in this more complex setting. With 300 pre-training queries, we now see all baselines generally outperform random initialization. 
This aligns with expectations since a larger pre-training budget gives multi-task methods more data per reward head, enabling better use of pre-training reward signals to learn meaningful representations.
\vspace{-0.5em}
\section{User Study}
Given the demonstrated benefits of our approach in simulation, we ask: can real humans label contextual feature queries to teach their personal contextual preferences? To investigate, we had participants label contextual feature queries and compare calibrated feature representations trained with varying query amounts, revealing whether increasing labeled input improves preference capture.

We conducted an in-person, IRB-approved study with 12 participants (7F/5M, aged 22-65, avg=29) recruited from the campus community, each paid \$25 for 70-90 minute sessions. Participants taught their contextual preferences for two \textit{Utensil} environment features, \textit{human} and \textit{stove}, 
with feature order counterbalanced and contexts discretized for visualization purposes. After instruction on contextual feature queries and a familiarization round (50 queries for \textit{laptop} feature in \textit{Cup} environment), participants completed the two experimental rounds (100 queries each). More details about the surveys and user interface can be found in App.~\ref{app:user-procedures} \&~\ref{app:user-guis}.

\para{Manipulated variable.} For each feature, we trained calibrated features using three different datasets: the first 25 queries, first 50, and all 100. Participants used a graphical user interface to inspect and evaluate four unlabeled models: these calibrated feature variants and the uncalibrated base feature. 

\para{Dependent variables.} Participants first wrote a brief description of their contextual preference, then responded to 100 queries with response times recorded. After query response, participants completed the System Usability Scale (SUS) and NASA-TLX. They then inspected and subjectively assessed the four presented models on a seven-point Likert scale, evaluating how well each captured their contextual preferences. The 4-item scale combined items adapted from Hoffman's goal subscale \cite{hoffman2019evaluating} (S1: \textit{The robot perceives accurately what my preferences are.}, S2: \textit{The robot does not understand how I would like the task to be completed.}), prior work in the assessment of adaptive agents \cite{nikolaidis2017human} (S3: \textit{The robot’s learned relationship is responsive to my input.}), and a task-specific question (S4: \textit{The robot’s learned relationship corresponds well with my expectations for how it should behave in different contexts.}). Participants then ranked the models by how well each captured their contextual preferences.

\para{Hypothesis.} Training with more user-labeled contextual feature queries improves alignment between learned calibrated features and the user's contextual preferences.
\begin{figure*}[t!]
    \centering    \includegraphics[width=0.85\textwidth, alt={Examples of two different user contextual preferences for the utensil feature, depicted by the user's written preference description the learned calibrated feature. Learned calibrated features are evaluated at all states in a point cloud and shown at different levels of utensil sharpness.}]{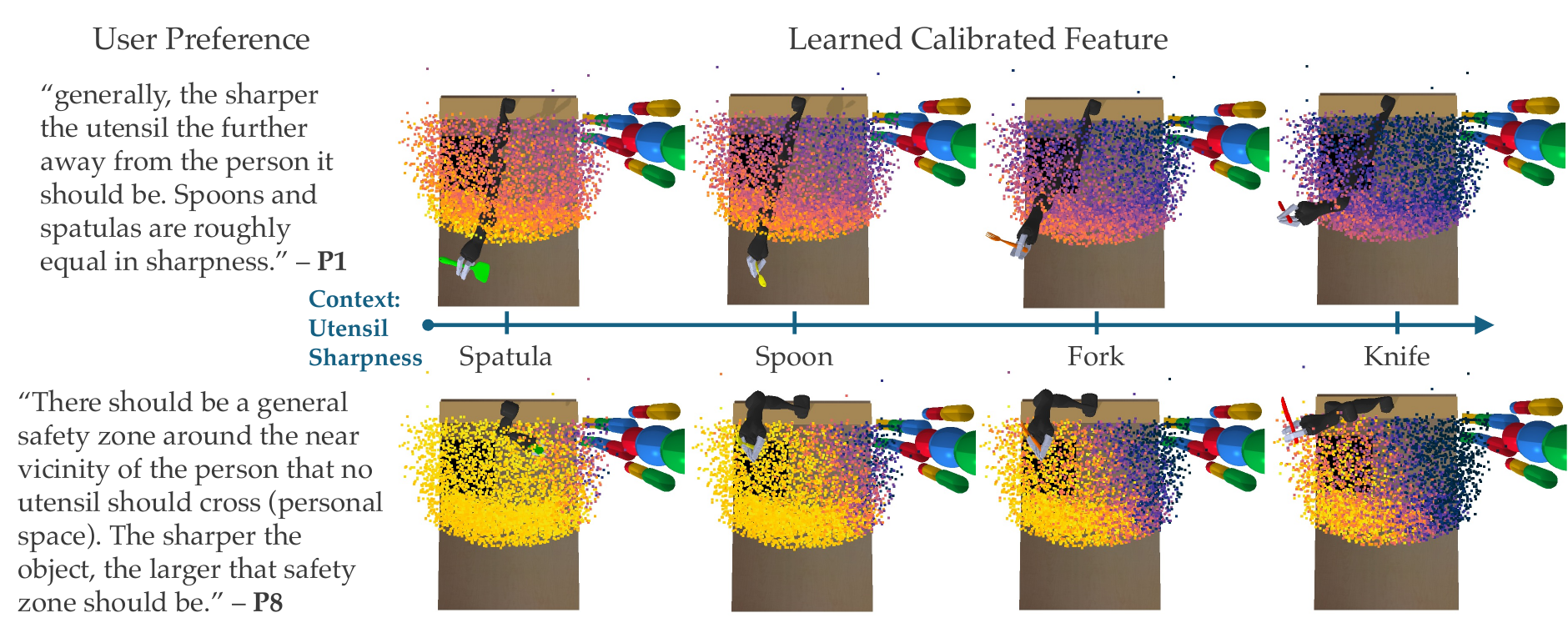}
    \caption{Examples of two unique user contextual preferences for how utensil sharpness should affect the \textit{human} feature (EE xy-planar distance to human). Users differ in their comfort with robot proximity and in which utensils they view as similar. By labeling contextual feature queries, users trained calibrated features that accurately captured these preferences.
    }
    \label{fig:participant-learned}
    \Description{Examples of two different user contextual preferences for the utensil feature, depicted by the user's written preference description the learned calibrated feature. Learned calibrated features are evaluated at all states in a point cloud and shown at different levels of utensil sharpness.}
    \vspace{-0.5em}
\end{figure*}
\vspace{-0.5em}
\subsection{Results}
The results predominantly support our hypothesis: both Likert ratings and average rankings improved with the number of contextual feature queries (Fig.~\ref{fig:hpe_data}). A two-way repeated measures ANOVA (feature type × query count) revealed a significant main effect of query count, $F(3, 33)=19.49, p<0.001, \eta^2_p=0.427$, but no effect of feature type, $F(1, 11)=0.09, p=0.77, \eta^2_p=0.0005$, and no interaction, $F(3, 33)=1.33, p=0.28, \eta^2_p=0.0333$. Bonferroni-corrected post-hoc tests showed significant differences in Likert ratings between all calibrated features and the base feature $(p < 0.05)$, and between 25 and 50 queries $(p < 0.05)$. Average Likert scores and rankings (1 = best, 4 = worst) improved with the number of training queries, though values leveled off between 50 and 100 queries.

Overall, calibrated features trained with any amount of contextual feature queries were significantly more aligned with user preferences than the base feature ($p<0.05$). Additionally, features trained with 50\% of queries showed significantly better alignment than those trained with 25\% ($p<0.05$). While further significance was not observed, likely due to variance across 12 participants, subjective ratings consistently increased with more queries. Thus, we find general support that increasing contextual feature queries improves alignment with user contextual preferences.

The user study additionally demonstrated the practical feasibility of the proposed method. 
The average response time per feature (100 queries) was 11.56 $\pm$ 2.36 minutes, indicating that contextual feature queries provide an efficient way to capture these personal contextual preferences. Moreover, Fig.~\ref{fig:hpe_data} shows different features need different amounts of data to converge in training, ranging from 50 to 100 queries in our experiments. The \textit{stove} feature showed similar Likert scores at 50 and 100 queries, indicating 50 queries sufficed. Participants also reported acceptable usability on the SUS (M = 71.0, SD = 18.3) and moderate workload on the NASA-TLX: mental (8.0 ± 5.1), physical (1.8 ± 1.9), temporal (9.6 ± 4.2), performance (13.8 ± 3.3), effort (8.0 ± 4.9), and frustration (4.1 ± 3.6), with both metrics averaged over the two query response rounds. 

Finally, post-hoc analysis of written user preference descriptions revealed diverse preferences, which we grouped into shared high-level categories (see App.~\ref{app:pref-analysis} for analysis details and all descriptions).
While we observe general trends of contextual preference, there is notable individual variation (Fig.~\ref{fig:participant-learned}). For the \textit{human} feature, 6/12 participants preferred a similar general preference while the remaining 6 described 5 different variations. For the \textit{stove} feature, 7/12 of participants preferred a similar general preference, while 5/12 held unique variants. These results suggest that while crowdsourcing or warm-starting general preferences may be useful, personalization is crucial to account for individual variations.
\vspace{-0.5em}
\section{Discussion and Limitations} \label{sec:discussion}
Our method faces several limitations. 
In continuous spaces, randomly sampled pairwise comparisons can miss important regions of the state space, which hurts our method's sample efficiency. Fig.~\ref{fig:calib_feats} shows this omission can prevent capturing feature irrelevance in specific contexts, such as when the stove is off or cup empty. It can also miss variations in small regions of the state. For example, when the cup is full, the calibrated \textit{laptop} feature collapses to low values across most states, although the point-cloud edge should remain high. 
These issues do not appear in the user study discrete contexts (Fig.~\ref{fig:participant-learned}). Future work could improve sample efficiency and coverage through active learning or dense data collection in key contexts, drawing on feature traces~\cite{bobu2022inducing}. Also, our post hoc analysis (App.~\ref{app:pref-analysis}) suggests users’ contextual preferences often share similarities, implying representations could be warm-started with crowdsourced data and refined via our method for necessary personalization.

Additionally, our method requires interpretable base features. Recent works use large-language or vision-language models to generate reward features as code automatically from raw state \cite{chenelemental, peng2024adaptive}, which are complementary approaches that could be used to generate a set of base reward features. In our user study we did not test whether users can identify which base features are contextually-affected, but their semantic meaning suggests this could be feasible in practice. Prior work shows that users naturally reference and correct individual features when behavior is inappropriate \cite{bajcsy2017learning}. Thus, we hypothesize that in deployment contextualized features can be obtained by either (1) asking users which features should vary with context, or (2) inferring the implicated feature from user corrections.

Finally, we consider practical limits to scaling to larger state spaces (e.g., images). Future work could explore composing state representations from image features (e.g., using DINOv3 \cite{simeoni2025dinov3}) and improving sample efficiency through active learning. When environments change (e.g., new objects are added), existing calibrated features remain valid unless the input state space changes. However, while new base features (e.g., distances to new objects) may share similar contextual adaptations (e.g., further EE distance depending on manipulated cup fullness), our current pipeline requires a new calibrated feature to be learned for each base feature. Future work could explore learning representations that generalize across object types, 
such as all ``technology" or all ``fragile" objects.

In summary, we show that explicit contextual modeling enables more efficient learning of robot rewards that adapt across contexts. By targeting data to teach contextual effects, our method learns useful representations from minimal human input. Its feature-based structure allows learned representations to be modularly composed into new rewards--a capability missing in baselines. 
 This structure also suggests representations have potential for interpretability. Future work could assess interpretability for supporting iterative refinement and appropriate trust calibration, both essential for responsible robot deployment. Our user study shows people can teach their contextual preferences, and the diversity of these preferences underscores the value of our method’s sample efficiency. By making it easier for users to teach complex behaviors, we take a step toward achieving preference alignment in widespread robot deployment.
\begin{acks}
We thank Mike Hagenow and Ho Chit Siu for helpful discussions. This work was partially funded by the Department of Defense (DoD) through the National Defense Science and Engineering Graduate (NDSEG) Fellowship Program.
\end{acks}
\bibliographystyle{ACM-Reference-Format}
\bibliography{HRI-2026}

@article{bobu2022inducing, title={Inducing structure in reward learning by learning features}, volume={41}, ISSN={1741-3176}, url={http://dx.doi.org/10.1177/02783649221078031}, DOI={10.1177/02783649221078031}, number={5}, journal={The International Journal of Robotics Research}, publisher={SAGE Publications}, author={Bobu, Andreea and Wiggert, Marius and Tomlin, Claire and Dragan, Anca D}, year={2022}, month=apr, pages={497–518}}

@misc{coumans2016pybullet,
  author       = {Erwin Coumans and Yunfei Bai},
  title        = {PyBullet, a Python module for physics simulation for games, robotics and machine learning},
  year         = {2016--2022},
  url          = {https://pybullet.org}
}

@inproceedings{abbeel2004apprenticeship, series={ICML ’04}, title={Apprenticeship learning via inverse reinforcement learning}, url={http://dx.doi.org/10.1145/1015330.1015430}, DOI={10.1145/1015330.1015430}, booktitle={Twenty-first international conference on Machine learning  - ICML ’04}, publisher={ACM Press}, author={Abbeel, Pieter and Ng, Andrew Y.}, year={2004}, pages={1}, collection={ICML ’04} }

@inproceedings{ziebart2008maximum,
  title={Maximum entropy inverse reinforcement learning},
  author={Ziebart, Brian D and Maas, Andrew L and Bagnell, J Andrew and Dey, Anind K and others},
  booktitle={AAAI},
  volume={8},
  pages={1433--1438},
  year={2008},
  organization={Chicago, IL, USA},
  url={https://cdn.aaai.org/AAAI/2008/AAAI08-227.pdf}
}

@inproceedings{christiano2017deep,
  author       = {Paul F. Christiano and
                  Jan Leike and
                  Tom B. Brown and
                  Miljan Martic and
                  Shane Legg and
                  Dario Amodei},
  title        = {Deep Reinforcement Learning from Human Preferences},
  booktitle    = {Advances in Neural Information Processing Systems 30: Annual Conference
                  on Neural Information Processing Systems 2017},
  pages        = {4299--4307},
  year         = {2017},
  url          = {https://proceedings.neurips.cc/paper/2017/hash/d5e2c0adad503c91f91df240d0cd4e49-Abstract.html},
  bibsource    = {dblp computer science bibliography, https://dblp.org}
}

@inproceedings{sadigh2017active, series={RSS2017}, title={Active Preference-Based Learning of Reward Functions}, url={http://dx.doi.org/10.15607/rss.2017.xiii.053}, DOI={10.15607/rss.2017.xiii.053}, booktitle={Robotics: Science and Systems XIII}, publisher={Robotics: Science and Systems Foundation}, author={Sadigh, Dorsa and Dragan, Anca and Sastry, Shankar and Seshia, Sanjit}, year={2017}, month=jul, collection={RSS2017} }

@inproceedings{bajcsy2017learning,
  author       = {Andrea Bajcsy and
                  Dylan P. Losey and
                  Marcia K. O'Malley and
                  Anca D. Dragan},
  title        = {Learning Robot Objectives from Physical Human Interaction},
  booktitle    = {1st Annual Conference on Robot Learning, CoRL 2017},
  pages        = {217--226},
  publisher    = {{PMLR}},
  year         = {2017},
  url          = {http://proceedings.mlr.press/v78/bajcsy17a.html}
}

@article{jain2015learning, title={Learning preferences for manipulation tasks from online coactive feedback}, volume={34}, ISSN={1741-3176}, url={http://dx.doi.org/10.1177/0278364915581193}, DOI={10.1177/0278364915581193}, number={10}, journal={The International Journal of Robotics Research}, publisher={SAGE Publications}, author={Jain, Ashesh and Sharma, Shikhar and Joachims, Thorsten and Saxena, Ashutosh}, year={2015}, month=may, pages={1296–1313} }

@inproceedings{bobu2023sirl, series={HRI ’23}, title={SIRL: Similarity-based Implicit Representation Learning}, url={http://dx.doi.org/10.1145/3568162.3576989}, DOI={10.1145/3568162.3576989}, booktitle={Proceedings of the 2023 ACM/IEEE International Conference on Human-Robot Interaction}, publisher={ACM}, author={Bobu, Andreea and Liu, Yi and Shah, Rohin and Brown, Daniel S. and Dragan, Anca D.}, year={2023}, month=mar, pages={565–574}, collection={HRI ’23} }

@inproceedings{yang2024enhancing, series={HRI ’24}, title={Enhancing Safety in Learning from Demonstration Algorithms via Control Barrier Function Shielding}, url={http://dx.doi.org/10.1145/3610977.3635002}, DOI={10.1145/3610977.3635002}, booktitle={Proceedings of the 2024 ACM/IEEE International Conference on Human-Robot Interaction}, publisher={ACM}, author={Yang, Yue and Chen, Letian and Zaidi, Zulfiqar and van Waveren, Sanne and Krishna, Arjun and Gombolay, Matthew}, year={2024}, month=mar, pages={820–829}, collection={HRI ’24} }

@inproceedings{ghosal2023contextual,
  author       = {Gaurav Rohit Ghosal and
                  Amrith Setlur and
                  Daniel S. Brown and
                  Anca D. Dragan and
                  Aditi Raghunathan},
  title        = {Contextual Reliability: When Different Features Matter in Different
                  Contexts},
  booktitle    = {International Conference on Machine Learning, {ICML} 2023},
  pages        = {11300--11320},
  publisher    = {{PMLR}},
  year         = {2023},
  url          = {https://proceedings.mlr.press/v202/ghosal23a.html}
}

@inproceedings{nishi2020fine, title={Fine-Grained Driving Behavior Prediction via Context-Aware Multi-Task Inverse Reinforcement Learning}, url={http://dx.doi.org/10.1109/icra40945.2020.9197126}, DOI={10.1109/icra40945.2020.9197126}, booktitle={2020 IEEE International Conference on Robotics and Automation (ICRA)}, publisher={IEEE}, author={Nishi, Kentaro and Shimosaka, Masamichi}, year={2020}, month=may, pages={2281–2287} }

@inproceedings{finn2017model,
  author       = {Chelsea Finn and
                  Pieter Abbeel and
                  Sergey Levine},
  title        = {Model-Agnostic Meta-Learning for Fast Adaptation of Deep Networks},
  booktitle    = {Proceedings of the 34th International Conference on Machine Learning,
                  {ICML} 2017},
  pages        = {1126--1135},
  publisher    = {{PMLR}},
  year         = {2017},
  url          = {http://proceedings.mlr.press/v70/finn17a.html}
}

@inproceedings{yu2019meta,
  author       = {Lantao Yu and
                  Tianhe Yu and
                  Chelsea Finn and
                  Stefano Ermon},
  title        = {Meta-Inverse Reinforcement Learning with Probabilistic Context Variables},
  booktitle    = {Advances in Neural Information Processing Systems 32: Annual Conference
                  on Neural Information Processing Systems 2019, NeurIPS 2019},
  pages        = {11749--11760},
  year         = {2019},
  url          = {https://proceedings.neurips.cc/paper/2019/hash/30de24287a6d8f07b37c716ad51623a7-Abstract.html}
}

@inproceedings{xu2019learning,
  author       = {Kelvin Xu and
                  Ellis Ratner and
                  Anca D. Dragan and
                  Sergey Levine and
                  Chelsea Finn},
  title        = {Learning a Prior over Intent via Meta-Inverse Reinforcement Learning},
  booktitle    = {Proceedings of the 36th International Conference on Machine Learning,
                  {ICML} 2019},
  pages        = {6952--6962},
  publisher    = {{PMLR}},
  year         = {2019},
  url          = {http://proceedings.mlr.press/v97/xu19d.html}
}

@inproceedings{seyed2019smile,
  author       = {Seyed Kamyar Seyed Ghasemipour and
                  Shixiang Gu and
                  Richard S. Zemel},
  title        = {SMILe: Scalable Meta Inverse Reinforcement Learning through Context-Conditional
                  Policies},
  booktitle    = {Advances in Neural Information Processing Systems 32: Annual Conference
                  on Neural Information Processing Systems 2019, NeurIPS 2019},
  pages        = {7879--7889},
  year         = {2019},
  url          = {https://proceedings.neurips.cc/paper/2019/hash/2b8f621e9244cea5007bac8f5d50e476-Abstract.html}
}

@article{gleave2018multi,
  author       = {Adam Gleave and
                  Oliver Habryka},
  title        = {Multi-task Maximum Entropy Inverse Reinforcement Learning},
  journal      = {CoRR},
  volume       = {abs/1805.08882},
  year         = {2018},
  url          = {https://doi.org/10.48550/arXiv.1805.08882},
  DOI = {10.48550/arXiv.1805.08882}
}

@inproceedings{chen2023multi,
  author       = {Jiayu Chen and
                  Dipesh Tamboli and
                  Tian Lan and
                  Vaneet Aggarwal},
  title        = {Multi-task Hierarchical Adversarial Inverse Reinforcement Learning},
  booktitle    = {International Conference on Machine Learning, {ICML} 2023},
  pages        = {4895--4920},
  publisher    = {{PMLR}},
  year         = {2023},
  url          = {https://proceedings.mlr.press/v202/chen23x.html}
}

@article{yamada2022task,
  author       = {Jun Yamada and
                  Karl Pertsch and
                  Anisha Gunjal and
                  Joseph J. Lim},
  title        = {Task-Induced Representation Learning},
  journal    = {The Tenth International Conference on Learning Representations, {ICLR}
                  2022},
  publisher    = {OpenReview.net},
  year         = {2022},
  url          = {https://openreview.net/forum?id=OzyXtIZAzFv}
}

@inproceedings{wang2021meta, title={Meta-Adversarial Inverse Reinforcement Learning for Decision-making Tasks}, url={http://dx.doi.org/10.1109/icra48506.2021.9561330}, DOI={10.1109/icra48506.2021.9561330}, booktitle={2021 IEEE International Conference on Robotics and Automation (ICRA)}, publisher={IEEE}, author={Wang, Pin and Li, Hanhan and Chan, Ching-Yao}, year={2021}, month=may, pages={12632–12638} }

@inproceedings{ng2000algorithms,
  author       = {Andrew Y. Ng and
                  Stuart Russell},
  title        = {Algorithms for Inverse Reinforcement Learning},
  booktitle    = {Proceedings of the Seventeenth International Conference on Machine
                  Learning {ICML} 2000},
  pages        = {663--670},
  publisher    = {Morgan Kaufmann},
  year         = {2000}
}

@inproceedings{zhao2022effectiveness,
  author       = {Mandi Zhao and
                  Pieter Abbeel and
                  Stephen James},
  title        = {On the Effectiveness of Fine-tuning Versus Meta-reinforcement Learning},
  booktitle    = {Advances in Neural Information Processing Systems 35: Annual Conference
                  on Neural Information Processing Systems 2022, NeurIPS 2022, New Orleans,
                  LA, USA, November 28 - December 9, 2022},
  year         = {2022},
  url          = {http://papers.nips.cc/paper\_files/paper/2022/hash/a951f595184aec1bb885ce165b47209a-Abstract-Conference.html}
}

@inproceedings{babes2011apprenticeship,
  author       = {Monica Babes and
                  Vukosi Marivate and
                  Kaushik Subramanian and
                  Michael L. Littman},
  title        = {Apprenticeship Learning About Multiple Intentions},
  booktitle    = {Proceedings of the 28th International Conference on Machine Learning,
                  {ICML} 2011},
  pages        = {897--904},
  publisher    = {Omnipress},
  year         = {2011},
  url          = {https://icml.cc/2011/papers/478\_icmlpaper.pdf}
}

@inproceedings{choi2012nonparametric,
  author       = {Jaedeug Choi and
                  Kee{-}Eung Kim},
  title        = {Nonparametric Bayesian Inverse Reinforcement Learning for Multiple
                  Reward Functions},
  booktitle    = {Advances in Neural Information Processing Systems 25: 26th Annual
                  Conference on Neural Information Processing Systems 2012.},
  pages        = {314--322},
  year         = {2012},
  url          = {https://proceedings.neurips.cc/paper/2012/hash/140f6969d5213fd0ece03148e62e461e-Abstract.html}
}

@inbook{dimitrakakis2012bayesian, title={Bayesian Multitask Inverse Reinforcement Learning}, ISBN={9783642299469}, ISSN={1611-3349}, url={http://dx.doi.org/10.1007/978-3-642-29946-9_27}, DOI={10.1007/978-3-642-29946-9_27}, booktitle={Recent Advances in Reinforcement Learning}, publisher={Springer Berlin Heidelberg}, author={Dimitrakakis, Christos and Rothkopf, Constantin A.}, year={2012}, pages={273–284} }

@inproceedings{finn2016gcl,
  author       = {Chelsea Finn and
                  Sergey Levine and
                  Pieter Abbeel},
  title        = {Guided Cost Learning: Deep Inverse Optimal Control via Policy Optimization},
  booktitle    = {Proceedings of the 33nd International Conference on Machine Learning,
                  {ICML} 2016},
  pages        = {49--58},
  publisher    = {JMLR.org},
  year         = {2016},
  url          = {http://proceedings.mlr.press/v48/finn16.html}
}

@article{bradley1952rank, title={Rank Analysis of Incomplete Block Designs: I. The Method of Paired Comparisons}, volume={39}, ISSN={0006-3444}, url={http://dx.doi.org/10.2307/2334029}, DOI={10.2307/2334029}, number={3/4}, journal={Biometrika}, publisher={JSTOR}, author={Bradley, Ralph Allan and Terry, Milton E.}, year={1952}, month=dec, pages={324} }

@inproceedings{sodhani2021multi,
  author       = {Shagun Sodhani and
                  Amy Zhang and
                  Joelle Pineau},
  title        = {Multi-Task Reinforcement Learning with Context-based Representations},
  booktitle    = {Proceedings of the 38th International Conference on Machine Learning,
                  {ICML} 2021},
  pages        = {9767--9779},
  publisher    = {{PMLR}},
  year         = {2021},
  url          = {http://proceedings.mlr.press/v139/sodhani21a.html}
}

@inproceedings{peng2024adaptive,
  author       = {Andi Peng and
                  Belinda Z. Li and
                  Ilia Sucholutsky and
                  Nishanth Kumar and
                  Julie Shah and
                  Jacob Andreas and
                  Andreea Bobu},
  title        = {Adaptive Language-Guided Abstraction from Contrastive Explanations},
  booktitle    = {Conference on Robot Learning},
  pages        = {3425--3438},
  publisher    = {{PMLR}},
  year         = {2024},
  url          = {https://proceedings.mlr.press/v270/peng25c.html}
}

@inproceedings{nikolaidis2017human, series={HRI ’17}, title={Human-Robot Mutual Adaptation in Shared Autonomy}, url={http://dx.doi.org/10.1145/2909824.3020252}, DOI={10.1145/2909824.3020252}, booktitle={Proceedings of the 2017 ACM/IEEE International Conference on Human-Robot Interaction}, publisher={ACM}, author={Nikolaidis, Stefanos and Zhu, Yu Xiang and Hsu, David and Srinivasa, Siddhartha}, year={2017}, month=mar, pages={294–302}, collection={HRI ’17} }

@article{o2021hierarchical, title={The Hierarchical Construction of Value}, url={http://dx.doi.org/10.31234/osf.io/9uvqp}, DOI={10.31234/osf.io/9uvqp}, publisher={Center for Open Science}, author={O’Doherty, John and Rutishauser, Ueli and Iigaya, Kiyohito}, year={2020}, month=dec }

@article{hoffman2019evaluating, title={Evaluating Fluency in Human–Robot Collaboration}, volume={49}, ISSN={2168-2305}, url={http://dx.doi.org/10.1109/thms.2019.2904558}, DOI={10.1109/thms.2019.2904558}, number={3}, journal={IEEE Transactions on Human-Machine Systems}, publisher={Institute of Electrical and Electronics Engineers (IEEE)}, author={Hoffman, Guy}, year={2019}, month=jun, pages={209–218} }

@inproceedings{mu2025preference, title={Preference-Based Multi-Objective Reinforcement Learning with Explicit Reward Modeling}, url={http://dx.doi.org/10.1109/cac63892.2024.10865310}, DOI={10.1109/cac63892.2024.10865310}, booktitle={2024 China Automation Congress (CAC)}, publisher={IEEE}, author={Mu, Ni and Luan, Yao and Jia, Qing-Shan}, year={2024}, month=nov, pages={4874–4879} }

@article{van2014multi,
  author       = {Kristof Van Moffaert and
                  Ann Now{\'{e}}},
  title        = {Multi-objective reinforcement learning using sets of pareto dominating policies},
  journal      = {The Journal of Machine Learning Research},
  pages        = {3483--3512},
  year         = {2014},
  url          = {https://dl.acm.org/doi/10.5555/2627435.2750356},
  doi          = {10.5555/2627435.2750356}
}

@inproceedings{chenelemental,
  author       = {Letian Chen and
                  Nina Marie Moorman and
                  Matthew Craig Gombolay},
  title        = {{ELEMENTAL:} Interactive Learning from Demonstrations and Vision-Language
                  Models for Reward Design in Robotics},
  booktitle    = {Forty-second International Conference on Machine Learning, {ICML}
                  2025},
  year         = {2025},
  url          = {https://openreview.net/forum?id=grlezgVg4s}
}

@article{simeoni2025dinov3,
  author       = {Oriane Sim{\'{e}}oni and
                  Huy V. Vo and
                  Maximilian Seitzer and
                  Federico Baldassarre and
                  Maxime Oquab and
                  Cijo Jose and
                  Vasil Khalidov and
                  Marc Szafraniec and
                  Seung Eun Yi and
                  Micha{\"{e}}l Ramamonjisoa and
                  Francisco Massa and
                  Daniel Haziza and
                  Luca Wehrstedt and
                  Jianyuan Wang and
                  Timoth{\'{e}}e Darcet and
                  Th{\'{e}}o Moutakanni and
                  Leonel Sentana and
                  Claire Roberts and
                  Andrea Vedaldi and
                  Jamie Tolan and
                  John Brandt and
                  Camille Couprie and
                  Julien Mairal and
                  Herv{\'{e}} J{\'{e}}gou and
                  Patrick Labatut and
                  Piotr Bojanowski},
  title        = {DINOv3},
  year         = {2025},
  url          = {https://doi.org/10.48550/arXiv.2508.10104},
  doi          = {10.48550/ARXIV.2508.10104},
  eprinttype    = {arXiv},
  eprint       = {2508.10104},
}

\pagebreak
\appendix
\section{Implementation details} \label{app:imp-details}
\begin{table*}[!h]
\centering
\caption{Representation learning hyperparameters.} \label{table:rep_params}
\begin{tabular}{lccccccc} 
\hline
\textbf{Model Type} & Optim. & LR & Batch Size & Weight Decay & $\lambda_{\text{reg}}$ & $\lambda_{FE/RE}$ & Epochs \\
\hline
Calibrated Features & Adam & $10^{-3}$ & 32 & 0.01 & $10^{-4}$ & 10 & 500 \\
\textit{Joint/SinglePref} & Adam & $10^{-3}$ & 32 & 0.01 & $10^{-4}$ & 10 & 3000 \\
\hline
\end{tabular}
\end{table*}

\begin{table*}[!h] 
\centering 
\caption{Reward learning hyperparameters.} \label{table:rew_params}
\begin{tabular}{lccccccc} 
\hline
\textbf{Model Type} & Optim. & LR & Batch Size & Weight Decay & $\lambda_{\text{reg}}$ & $\lambda_{FE/RE}$ & Epochs \\
\hline
Calibrated Features & Adam & $10^{-2}$ & 32 & 0 & 0 & - & 200 \\
\textit{Joint}/\textit{SinglePref} & Adam & $10^{-4}$ & 64 & $10^{-3}$ & $10^{-3}$ & 1 & 500 \\
\textit{RandomInit} & Adam & $10^{-3}$ & 32 & $10^{-2}$ & $0$ & 1 & 500 \\
\hline
\end{tabular}
\end{table*}
\subsection{Loss regularization}
We implemented the paired comparisons loss (Eq.~\ref{eq:final-loss}\&~\ref{eq:final-loss-rew}) with an additional regularization term that penalizes the size of the model logits. For training the reward function $R_{\theta}$, this term regularization is,
\begin{equation}
\ell_{\text{reg}}(\theta; \mathcal{D}) = \sum_{(s_1, s_2, \ell) \in \mathcal{D}} (R_{\theta}(s_1)^2 + R_{\theta}(s_2)^2).
\end{equation}
Similarly, for training calibrated feature function $\phi^c_{\psi_i}$, the analogous regularization term is,
\begin{equation}
\ell_{\text{reg}}(\psi_i; \mathcal{D}) = \sum_{(s_1, s_2, \ell) \in \mathcal{D}} (\phi^c_{\psi_i}(s_1)^2 + \phi^c_{\psi_i}(s_2)^2).
\end{equation}
With this additional regularization, the implemented loss for training the reward function becomes,
\begin{equation}
\mathcal{L}(\theta) =
\frac{1}{\lvert D_{\theta}\rvert} \left(\lambda_{\text{RE}}\cdot \ell(\theta; D_{\theta}^{\text{equiv}}) + \ell(\theta; D_{\theta}^{\text{pref}}) + \lambda_{\text{reg}}\cdot \ell_{\text{reg}}(\theta; D_{\theta})\right).
\end{equation}
Likewise, for training each calibrated feature, the loss function is structured identically,
\begin{equation}
\mathcal{L}(\psi_i) =
\frac{1}{\vert D_{\psi_i}\rvert} \left(\lambda_{\text{FE}}\cdot \ell(\psi_i; D_{\psi_i}^{\text{equiv}}) + \ell(\psi_i; D_{\psi_i}^{\text{pref}}) + \lambda_{\text{reg}}\cdot \ell_{\text{reg}}(\psi_i; D_{\psi_i})\right).
\end{equation}
In both cases, $ \lambda_{\text{reg}}$ represents the same hyperparameter that weights the regularization term relative to the other loss terms.

\subsection{Model architectures and hyperparameters}
Here we present the selected architecture and hyperparameters and describe the selection process. For training stability, model inputs (state and features) were normalized to the $[0,1]$ range across the full 10k-state dataset in an environment.
\subsubsection{Selection} 
Architectures and hyperparameters were optimized separately for each representation learning approach. For each method, we first tuned the hyperparameters for representation learning, then tuned the reward hyperparameters on top of frozen learned representations. Reported hyperparameters were tuned, while all others were left as their PyTorch and optimizer defaults. We conducted all tuning experiments in the \textit{Weighted Block} environment. 

For representation learning methods, in order to maintain sample efficiency, we limited our evaluation to relatively simple MLP architectures, ranging from 1-3 layers and 32-128 units. We selected calibrated feature architecture and hyperparameters using model mean squared error (MSE), and multi-task architecture hyperparameters based on average reward accuracy across the training rewards. Hyperparameter tuning was performed exclusively on the multi-task model with 50 training preferences (\textit{JointPref} (50)) to leverage the most diverse representation learning scenario, and the resulting configuration was applied consistently across multi-task variants. 

For reward learning, we learned a linear combination of calibrated features for our method. For the multi-task baseline, we tested several different architectures, both linear combinations and the same simple MLP architectures used for representations. We gave the baseline access to more expressive reward head architectures than ours to provide the best possible advantage. Architecture and hyperparameters were selected based on reward accuracy across 10 test rewards.

For both representation and reward learning methods, we evaluated whether including or excluding equivalence-labeled queries improved performance. When excluded, equivalent queries were skipped and a new query asked. This strategy only improved performance when learning a linear reward over calibrated features.

\subsubsection{Representation learning (pre-training)}
Representation learning hyperparameters are summarized in Table~\ref{table:rep_params}. The \textit{JointPref} and \textit{SinglePref}   models used MLPs with three 32-unit layers, outputting to a 7-dimensional latent space. Each calibrated feature network also used three 32-unit layers, with a scalar output passed through a softmax output activation function to produce values in $\mathbb{R}^+$. Learned calibrated feature output is normalized using the maximum and minimum logits encountered during training, scaling that the calibrated feature output within the range $[0,1]$. For both approaches, we found that using LeakyReLU activations and Xavier initialization (scaled for LeakyReLU) improved training stability. During representation learning, training task reward heads for all multi-task baselines are modeled as linear combinations of latent representations in order to encourage learned representations to compactly capture meaningful task-relevant variation while remaining amenable to simple downstream reward inference.

\subsubsection{Reward learning}
Reward learning hyperparameters are listed in Table~\ref{table:rew_params}. For all methods we evaluated freezing or fine-tuning pre-trained representations, with details in App.~\ref{app:frozen-results}.
For our method, we learned a linear reward over calibrated features. Our method performed the best skipping equivalence queries and generating new non-equivalent queries instead, referenced by the null $\lambda_{\text{FE}}$ value. All multi-task variants used a reward network consisting of a single 32-unit MLP layer (with PyTorch default weight initialization and standard ReLU activations). For the  \textit{RandomInit} baseline, we tested learning rewards from scratch using the same end-to-end architecture as \textit{JointPref}, and with no compressed embedding, finding the same architecture as \textit{JointPref} to perform the best. We performed a separate round of hyperparameter tuning for \textit{RandomInit}, with selected hyperparameters also in Table~\ref{table:rew_params}. 

\subsection{Compute resources}
Each experiment was run on 8 CPU-only compute nodes, each equipped with Intel Xeon Platinum 8260 processors (48 cores per node) and 192 GB of RAM (approximately 4 GB per core). On each node, 24 parallel jobs were run using 2 cores per job. No GPUs were used. Most experiments--such as training a multi-task representation across six random seeds for each query count interval--completed within 5-7 minutes using this setup. For the user study, model training and evaluation were conducted locally on a laptop equipped with an Apple M1 chip and 16GB of memory. Training the four small models on this machine for each query response round of the experiment took approximately one minute.

\begin{figure*}[!h]
    \centering\includegraphics[width=0.8\linewidth, alt={Three manipulation environments. The robot is on a table near a stove and a laptop, and a human is standing nearby. The robot holds either a block, a cup, or a utensil. Features are visually depicted using blue arrows and labeled with text.}]{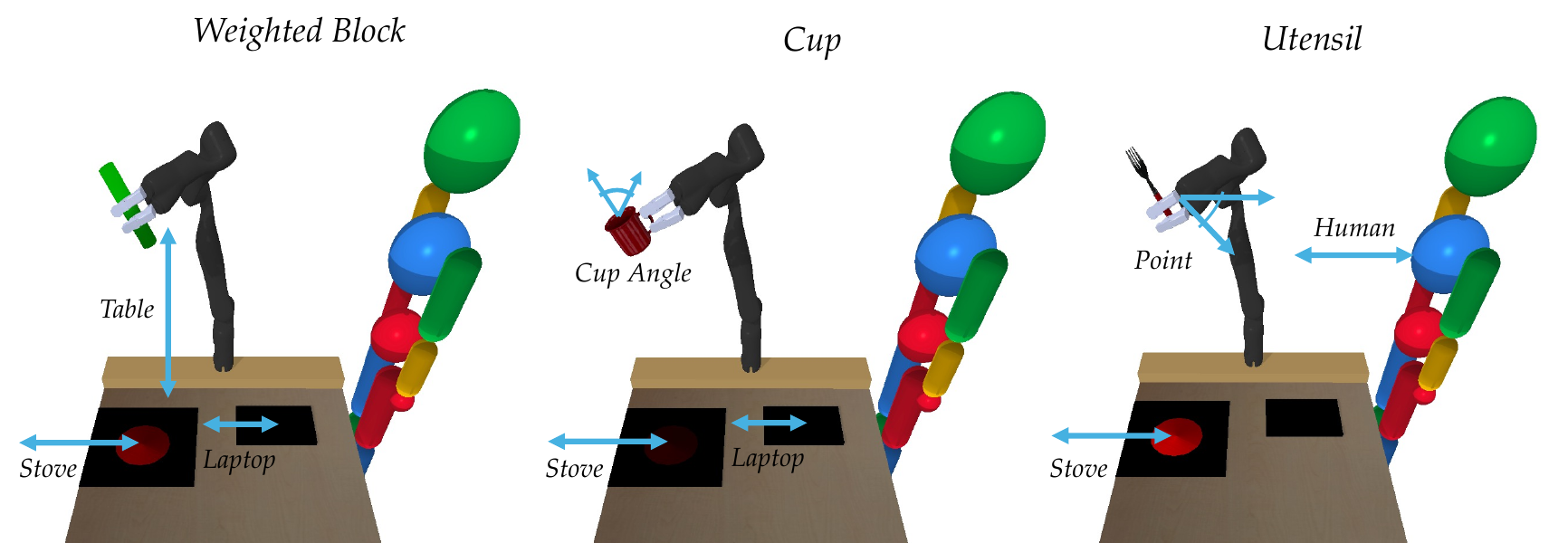}
    \caption{Three experimental environments with features shown in blue.}
    \label{fig:env-feats}
    \Description{Three manipulation environments. The robot is on a table near a stove and a laptop, and a human is standing nearby. The robot holds either a block, a cup, or a utensil. Features are visually depicted using blue arrows and labeled with text.}
\end{figure*}

\section{Environment details} \label{app:envs}
We generate 10k random states per environment by selecting a random robot EE xyz location above the table within the robot's reachable workspace, then solving inverse kinematics to obtain joint angles without collisions. 
Object positions (human, stove, and laptop) remain constant across states. The two contextual elements of a state are randomly sampled with continuous values that range between 0 and 1; however, our method can work with contextual elements that follow the state initialization generally in $\mathbb{R}$.

\subsection{Base feature definition}\label{app:base-feat-def} Fig.~\ref{fig:env-feats} depicts the three base features in each manipulation environment. The six unique base features definitions are detailed below: 
\begin{itemize}
    \item \textit{Table} measures the vertical distance between the EE and the table surface:
\begin{equation}
f_{\text{table}}(s) = p^z_{\text{EE}} - p^z_{\text{table}},
\end{equation}
where $p^z_{\text{EE}}$ is the z-coordinate of the EE position and $p^z_{\text{table}}$ is the z-coordinate of the table surface, which is set as a constant and not part of the state space. 
\item \textit{Laptop} measures the planar (XY) distance between the EE and the laptop, capped at a maximum value,
\begin{equation}
f_{\text{laptop}}(s) = \min(\|p^{\text{XY}}_{\text{EE}} - p^{\text{XY}}_{\text{laptop}}\|_2, 0.8),
\end{equation}
where $p^{\text{XY}}_{\text{EE}}$ denotes the XY-coordinates of the EE and $p^{\text{XY}}_{\text{laptop}}$ denotes the XY-coordinates of the laptop.
\item \textit{Human} measures the planar (XY) distance between the EE and the human, capped at a maximum value,
\begin{equation}
f_{\text{human}}(s) = \min(\|p^{\text{XY}}_{\text{EE}} - p^{\text{XY}}_{\text{human}}\|_2, 1.0)
\end{equation}
where $p^{\text{XY}}_{\text{human}}$ denotes the XY-coordinates of the human.
\item \textit{Point} measures how much a utensil held by the robot points toward the human,
\begin{equation}
f_{\text{point}}(s) = \vec{u}^{W}_{\text{Ex}}{}^\top \vec{u}^{W}_{\text{EH}}
\end{equation}
where
\begin{align}
\vec{u}^{,W}_{\text{Ex}} = R^{W}_{\text{EE}} \begin{bmatrix}1, \ 0, \ 0\end{bmatrix}^\top,
\quad \vec{u}^{,W}_{\text{EH}} = \frac{p^{W}_{\text{human}} - p^{W}_{\text{EE}}}{\left\|p^{W}_{\text{human}} - p^{W}_{\text{EE}}\right\|_2}.
\end{align}
Here, $\vec{u}^{,W}_{\text{Ex}}$ is the x-axis of the end-effector (i.e., the utensil’s "pointing" direction) expressed in the world frame, obtained via the rotation matrix $R^{W}_{\text{EE}}$ from the end-effector to the world frame. The term $\vec{u}^{W}_{\text{EH}}$ describes the unit vector from the end-effector to the human, also in the world frame. The function equals $1$ when the utensil points directly at the human, and $-1$ when it points directly away.
\item \textit{Cup angle} The coffee feature measures how upright the cup is held:
\begin{equation}
f_{\text{coffee}}(s) = \left[R^W_{WE}\right]_{3,1}
\end{equation}
where $\left[R^W_{WE}\right]_{3,1}$ denotes the element in the third row and first column of the rotation matrix from the EE frame to the world frame. This represents the z-component of the EE's x-axis in world coordinates. Since the cup's upright direction is aligned with the EE's x-axis, when this x-axis aligns with the world z-axis (vertical), the \textit{cup angle} value approaches 1, indicating maximum "uprightness".

\item \textit{Stove} The stove feature measures the planar (XY) distance between the EE and the stove, capped at a maximum value:
\begin{equation}
f_{\text{stove}}(s) = \min(\|p^{\text{XY}}_{\text{EE}} - p^{\text{XY}}_{\text{stove}}\|_2, 0.8)
\end{equation}
where $p^{\text{XY}}_{\text{stove}}$ denotes the XY-coordinates of the stove.
\end{itemize}

\subsection{Discrete contexts for visualization.} \label{sec:discrete-context} While all simulated human participant experiments use continuous contexts, visualization requires establishing a set of discrete values for each contextual element (such as a discrete set of utensils varying in sharpness). Each contextual element is discretized with different granularities, all evenly spaced between 0 and 1: stove heat (8 discrete values), block weight (16 discrete values), cup fullness (6 values), and utensil sharpness (4 values). These discrete values are used only for the user study and generating visualization states for qualitative feature visualizations.

\section{Simulated Experiments}
\subsection{Human simulation}

\subsubsection{Equivalence threshold selection}\label{app:sim_human-threshold}
To simulate human judgments of equivalence in paired comparisons, we introduce a threshold $\epsilon$ that defines when two states are considered equivalent. Specifically, this threshold determines whether $\lvert f(s_1) - f(s_2)\rvert \leq \epsilon$, in which case the simulated human labels the pair as equivalent, i.e., $f(s_1) = f(s_2)$. The function $f$ represents any operation on the state, such as a calibrated feature value or the reward. Using a common $\epsilon$ across both types is reasonable, since the ground truth reward is simply a linear combination (with  weights with maximum magnitude of 1), of three calibrated features (each with maximum magnitude of 1).

\begin{figure*}[t!]
    \centering
    \includegraphics[width=0.8\linewidth, alt={Line plots showing calibrated feature MSE against number of representation learning queries. Three lines shown for three different equivalence thresholds (0.01, 0.001, 0.0001). Line plots show minimal difference between the three lines.}]{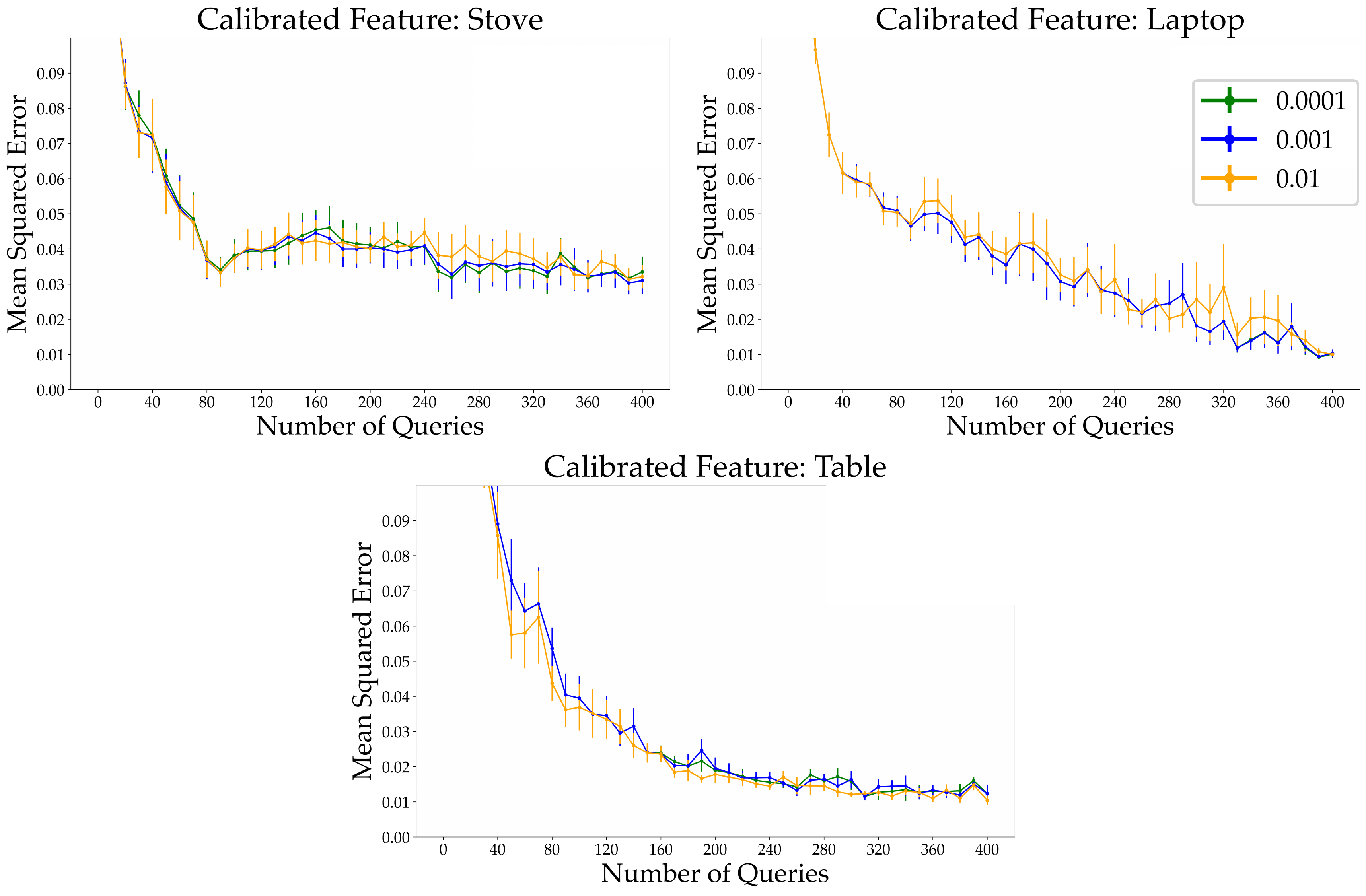}
    \caption{Equivalence threshold experiments for calibrated features.}
    \label{fig:threshold-cns}
    \Description{Line plots showing calibrated feature MSE against number of representation learning queries. Three lines shown for three different equivalence thresholds (0.01, 0.001, 0.0001). Line plots show minimal difference between the three lines.}
\end{figure*}

\begin{figure*}[!h]
    \centering
    \includegraphics[width=0.8\linewidth, alt={Line plots showing multi-task training rewards against number of representation learning queries. Three lines shown in each plot for three different equivalence thresholds (0.01, 0.001, 0.0001). One plot shown for each multi-task variant. Line plots show minimal difference between the three lines in each plot.}]{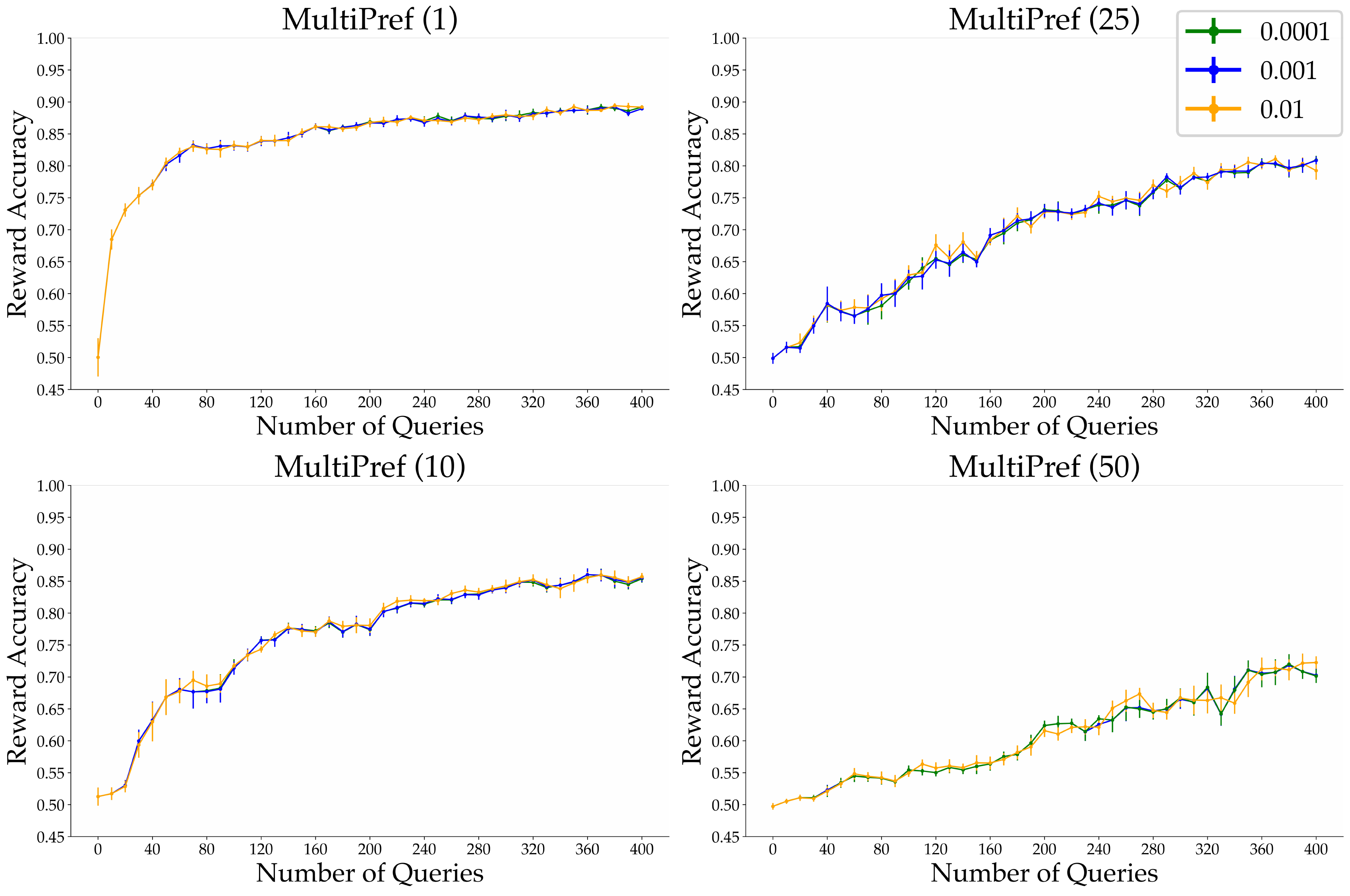}
    \caption{Equivalence threshold experiments for multi-task baselines.}
    \label{fig:threshold-multitask}
    \Description{Line plots showing multi-task training rewards against number of representation learning queries. Three lines shown in each plot for three different equivalence thresholds (0.01, 0.001, 0.0001). One plot shown for each multi-task variant. Line plots show minimal difference between the three lines in each plot.}
    \vspace{2em}
\end{figure*}

We select $\epsilon$ by evaluating representation learning under three different candidate thresholds in the \textit{Weighted Block} environment with all features calibrated. For our method, we plot the MSE between the learned and ground truth calibrated feature values (Fig~\ref{fig:threshold-cns}). For the \textit{Multipref} models, we plot the Reward Accuracy metric evaluated and averaged across all training rewards (Fig.~\ref{fig:threshold-multitask}).

We observe nearly identical trends across the three tested thresholds, including at key points where the learned representations are used for downstream tasks (i.e., at 100 queries per calibrated feature and 100/300 queries for each multi-task baseline). Based on this, we select $\epsilon = 0.01$, which is the coarsest of the three thresholds and provides the most realistic approximation of human equivalence judgments.

\begin{figure*}[h!]
    \centering    \includegraphics[width=0.70\linewidth, alt={3D figures showing how calibrated feature value is affected by changes in base feature and relevant contextual element.}]{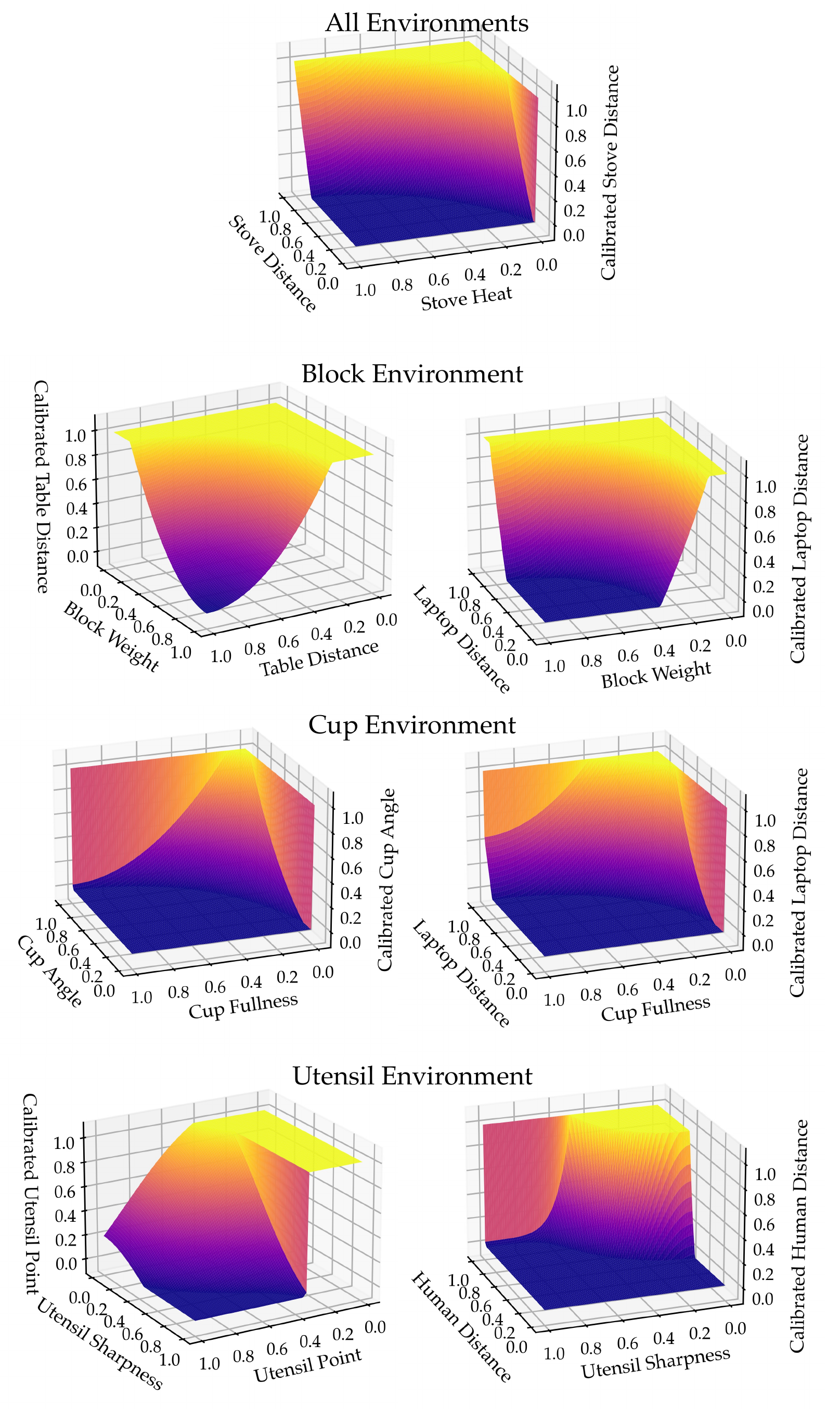}
    \caption{Ground truth calibrated feature functions with respect to base feature and affecting contextual element.}
    \label{fig:calib-feats-fns}
    \Description{3D figures showing how calibrated feature value is affected by changes in base feature and relevant contextual element.}
\end{figure*}

\subsubsection{Ground truth calibrated feature function definitions} \label{app:gt-calib-feat-def}
Ground truth calibrated feature functions were defined to represent complex three-dimensional relationships governing how context input should shape base features. Intuitively, picking a static context represents taking a slice of through a defined calibrated feature function, demonstrating how the calibrated feature varies with the input base feature under that context. We visualize the ground truth calibrated features as a function of the base feature and affecting contextual element in Fig.~\ref{fig:calib-feats-fns}. The actual functions being learned by calibrated feature networks are even more complex, given they must not only learn the visualized relationships, but additionally invariance to all other non-relevant dimensions of the state.

We define ground truth calibrated features such that they are between 0 and 1 under expected input conditions, when contextual element $c$ and input feature value $\phi$ are in their expected ranges, $c \in [0,1], \phi \in [0,1]$. Importantly, we define ground truth calibrated feature functions to describe how a single contextual element of the state affects a base feature; however, our method can handle any subset of the state contextually affecting a base feature, not just a single element.

\para{Base functions.} To help define calibrated feature functions, we first define a handful of base functions that are used to compose the ground truth calibrated feature functions:

\begin{enumerate}
    \item Gaussian function:
\begin{equation}
G(\phi, c; \sigma, \mu_\phi, \mu_c, N, b) = \frac{N}{2\pi\sigma^2}\exp\left(-\frac{(\phi-\mu_\phi)^2+(c-\mu_c)^2}{2\sigma^2}\right) + b, 
\end{equation}
which describes a bell-shaped surface with peak at $(\mu_\phi, \mu_c)$, decaying exponentially in all directions. Height controlled by $N$, spread by $\sigma$, and vertical shift by $b$.

\item Logistic function:
\begin{equation}
L(\phi; L, k, \mu_\phi, b) = \frac{L}{1+e^{-k(\phi-\mu_\phi)}} + b,
\end{equation}
which describes an S-shaped curve centered at $\mu_\phi$, with steepness $k$, maximum value $L$, and vertical shift $b$.

\item Bowl function:
\begin{equation}
B(\phi, c; N_\phi, N_c, \mu_\phi, \mu_c, b) = N_\phi(\phi-\mu_\phi)^2 + N_c(c-\mu_c)^2 + b,
\end{equation}
which generates a parabolic bowl with minimum at $(\mu_\phi, \mu_c)$, stretching factors $N_\phi$ and $N_c$, and vertical shift $b$.

\item Modified logistic:
\begin{equation}
X(\phi; L, k, \mu_\phi, b) = \frac{L}{1+e^{-k}(\phi-\mu_\phi)} + b,
\end{equation}
which produces a modified logistic function with denominator scaling, creating asymmetric S-curve.
\end{enumerate}

\para{Calibrated features.} All composed functions follow the form:
\begin{equation*}
f_i(\phi, c) = \min(\max(f_{\text{base}}(\phi, c), 0), 1)
\end{equation*}
where $\phi$ is the input base feature, $c$ is the relevant contextual state element, and $f_{\text{base}}$ is a single base function or some combination of them. The calibrated features are defined in this form as follows: 

\begin{enumerate}
    \item Calibrated \textit{stove} feature:
\begin{equation}
f_{\text{stove}}(\phi, c) = 
\begin{cases}
1 & \text{if}\ c = 0 \\
\min\Bigl(\max(B(\phi, c; 1, 2, -1, 1, -3), 0), 1\Bigr) & \text{if}\ c > 0
\end{cases}
\end{equation}

\item Calibrated \textit{table} feature:
\begin{equation}
f_{\text{table}}(\phi, c) = \min(B(\phi, c; 1.8, 1.5, 1, 1, 0), 1)
\end{equation}

\item Calibrated \textit{laptop} feature (as affected by block weight):
\begin{equation}
f_{\text{laptop (weight)}}(\phi, c) = 
\begin{cases}
1 & \text{if } c = 0 \text{ or } \phi \geq 1 \\[6pt]
  \min\Bigl( 
     \max\bigl( \\
  \hspace{0.5em}
     B(\phi, c; 2.5, 2.5, 0, 1, -1), 0 \bigr), 1 \Bigr)
& \text{otherwise}
\end{cases}
\end{equation}

\item Calibrated \textit{cup angle} feature:
\begin{equation}
f_{\text{cup angle}}(\phi, c) = 
\begin{cases}
1 & \text{if } c = 0\ \text{or}\ \phi \geq 1 \\
\min\Bigl(\max\Bigl( \\ \hspace{0.5em}
X(\phi; -2, -1.1, 2, -0.65) \\ \hspace{0.5em} + B(\phi, c; 0.5, 1.2, -0.2, 1, -1)
, 0\Bigr), 1\Bigr) & \text{otherwise}
\end{cases}
\end{equation}

\item Calibrated \textit{laptop} feature (as affected by cup fullness):
\begin{equation}
f_{\text{laptop (fullness)}}(\phi, c) = 
\begin{cases}
1 & \text{if}\ c = 0\ \text{or}\ \phi \geq 1 \\
\min\Bigl(\max(\\ \hspace{0.5em} B(\phi, c; 2, 1.5, 0, 1, -1.5), 0), 1\Bigr) & \text{otherwise}
\end{cases}
\end{equation}

\item Calibrated \textit{human} feature:
\begin{equation}
f_{\text{human}}(\phi, c) = 
\begin{cases}
1 & \text{if } \phi \geq 1 \\
\min\Bigl(\max\Bigl( \\ \hspace{0.5em}
G(\phi, c; 0.2, 1, 0, 6, -0.55) \\ 
\hspace{0.5em} + X(\phi; -2, -1.1, 2, -0.4)
, 0\Bigr), 1\Bigr) & \text{otherwise}
\end{cases}
\end{equation}

\item Calibrated \textit{point} feature:
\begin{equation}
f_{\text{point}}(\phi, c) = 
\begin{cases}
1 & \text{if } \phi < \frac{1}{3} \\
\min\Bigl(\max\Bigl( \\ \hspace{0.5em}
X(\phi; 0.2, -1, 0.5, 0) \\ \hspace{0.5em} + G(\phi, c; 0.5, 0.4, 0, 2, -0.5)
, 0\Bigr), 1\Bigr) & \text{otherwise.}
\end{cases}
\end{equation}
\end{enumerate}

\subsection{Representation learning results} \label{app:rep-learning}
\begin{figure*}[!t]
    \centering
    \includegraphics[width=0.8\linewidth, alt={Line plots showing calibrated feature MSE against number of representation learning queries. A separate plot is shown for each environment. MSE generally converges around 100 queries, except for the distance to human feature in the Utensil environment, which takes closer to 300  queries to converge.}]{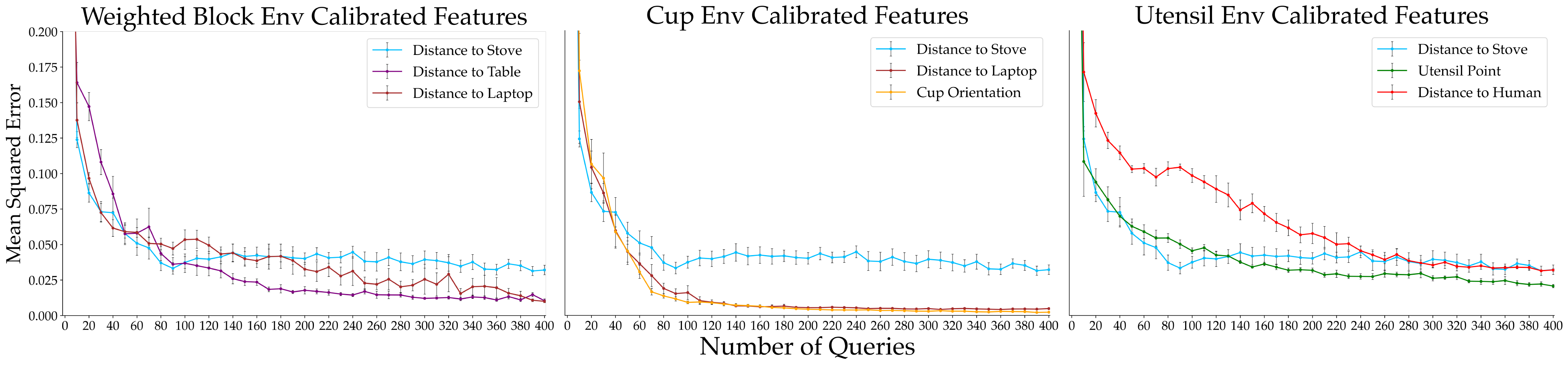}
    \caption{Calibrated feature learning results.}
    \label{fig:rep-cns}
    \Description{Line plots showing calibrated feature MSE against number of representation learning queries. A separate plot is shown for each environment. MSE generally converges around 100 queries, except for the distance to human feature in the Utensil environment, which takes closer to 300  queries to converge.}
\end{figure*}

\begin{figure*}[!t]
    \centering
    \includegraphics[width=0.7\linewidth, alt={Line plots showing training reward accuracy against number of representation learning queries, which in this case are training preference queries. Training reward accuracy generally takes longer to converge the greater the number of training rewards there are.}]{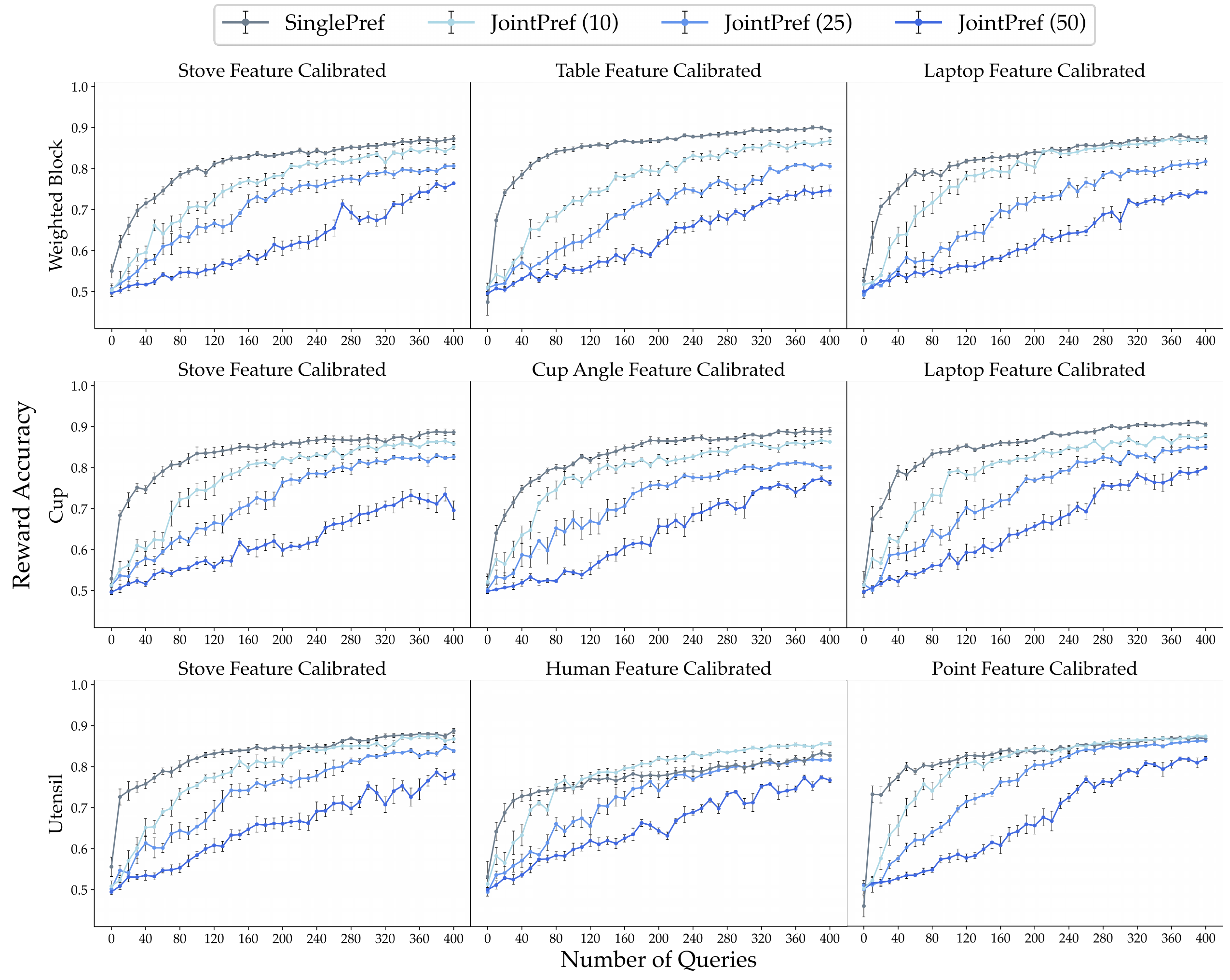}
    \caption{Multi-task representation learning results for single contextually-affected feature experiments.}
    \label{fig:rep-multitask-single-calib}
    \Description{Line plots showing training reward accuracy against number of representation learning queries, which in this case are training preference queries. Training reward accuracy generally takes longer to converge the greater the number of training rewards there are.}
\end{figure*}

\begin{figure*}
    \centering
    \includegraphics[width=0.75\linewidth, alt={Line plots showing training reward accuracy against number of representation learning queries, which in this case are training preference queries. Training reward accuracy generally takes longer to converge the greater the number of training rewards there are.}]{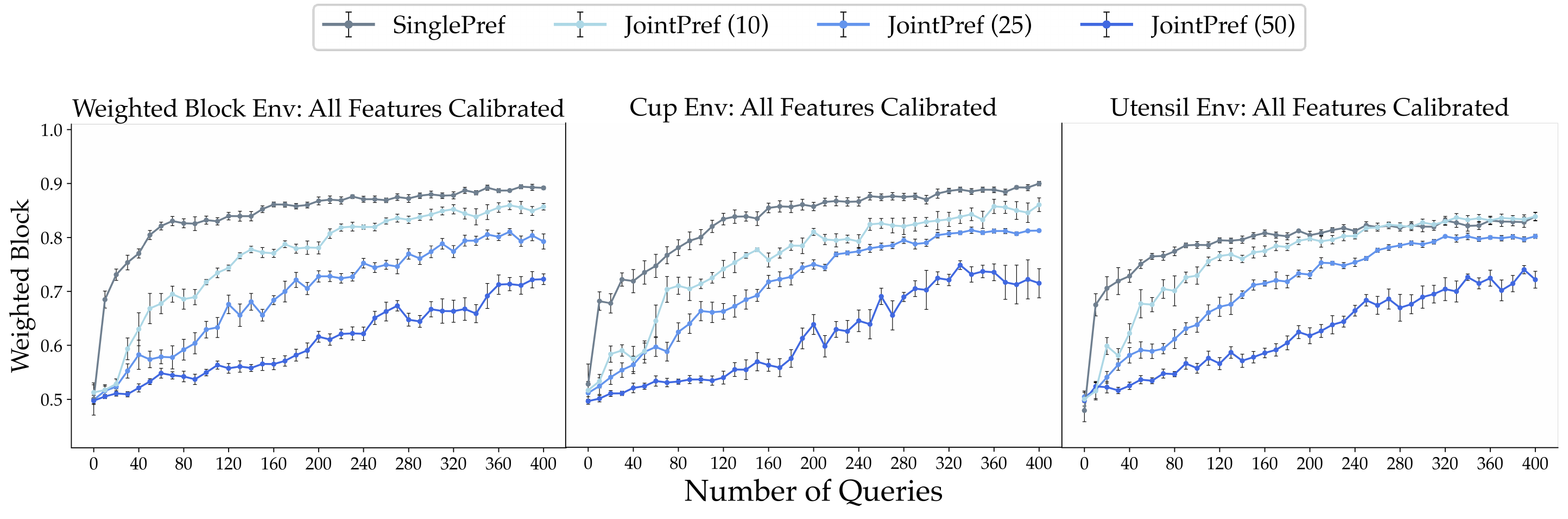}
    \caption{Multi-task representation learning results for multiple contextually-affected feature experiments.}
    \label{fig:rep-multitask-all-calib}
    \Description{Line plots showing training reward accuracy against number of representation learning queries, which in this case are training preference queries. Training reward accuracy generally takes longer to converge the greater the number of training rewards there are.}
\end{figure*}

We present intermediate representation learning results for our method and multi-task baselines (Fig.~\ref{fig:rep-cns}-~\ref{fig:rep-multitask-all-calib}). These visualizations were used to select the number of pre-training queries set for downstream reward learning. We select 100 pre-training queries given we see general convergence behavior for calibrated features at this point (Fig.~\ref{fig:rep-cns}). When all three features are contextually-affected we use 300 pre-training queries so that all calibrated features still see 100 pre-training queries when divided among the three features.

For calibrated features, we plot the mean squared error (MSE) of all three learned calibrated features for each environment as a function of the number of training queries (Fig.~\ref{fig:rep-cns}), evaluated across the 2k test states per seed. Calibrated feature MSE can struggle to approach zero when learning from comparisons, which only provide ordinal relationships between pairs of points rather than absolute function values. Some smoother functions (such as \textit{cup} and \textit{laptop}, which have similar defined calibrated feature functions) are easier to capture accurately with only pairwise comparisons, while others (e.g., \textit{human}) can struggle to achieve low MSE.

We next present multi-task representation learning results, which report reward accuracy averaged over the $N$ training rewards in each environment (Fig.~\ref{fig:rep-multitask-single-calib}\&~\ref{fig:rep-multitask-all-calib}). We can see that as expected, the \textit{SinglePref} model achieves high reward accuracy the most rapidly, given it is only trained and evaluated on a single reward. In contrast, as the number of $N$ training tasks for each \textit{JointPref} method increases, the more queries it takes to find a shared representation that supports accurate rewards across all training tasks. Notably, for multi-task baselines, representation learning was required to be repeated for the scenario in which all features are contextually-affected. This is because the contextual adaptations learned in each single contextually-affected feature scenario operate on the full feature set, therefore individual feature adaptations cannot be easily extracted and composed. In contrast, our calibrated feature approach supports reuse: the same three calibrated features were combined in the all calibrated features setting without retraining the representation.

\subsection{Freezing vs. fine-tuning representations} \label{app:frozen-results}
\begin{figure*}[!t]
    \centering
    \includegraphics[width=0.8\linewidth, alt={Line plots showing reward accuracy against number of reward preference queries. For each method a there are two lines: one with frozen representations and one with unfrozen representations. Our method performs better with frozen representations while baselines generally perform better with unfrozen representations.}]{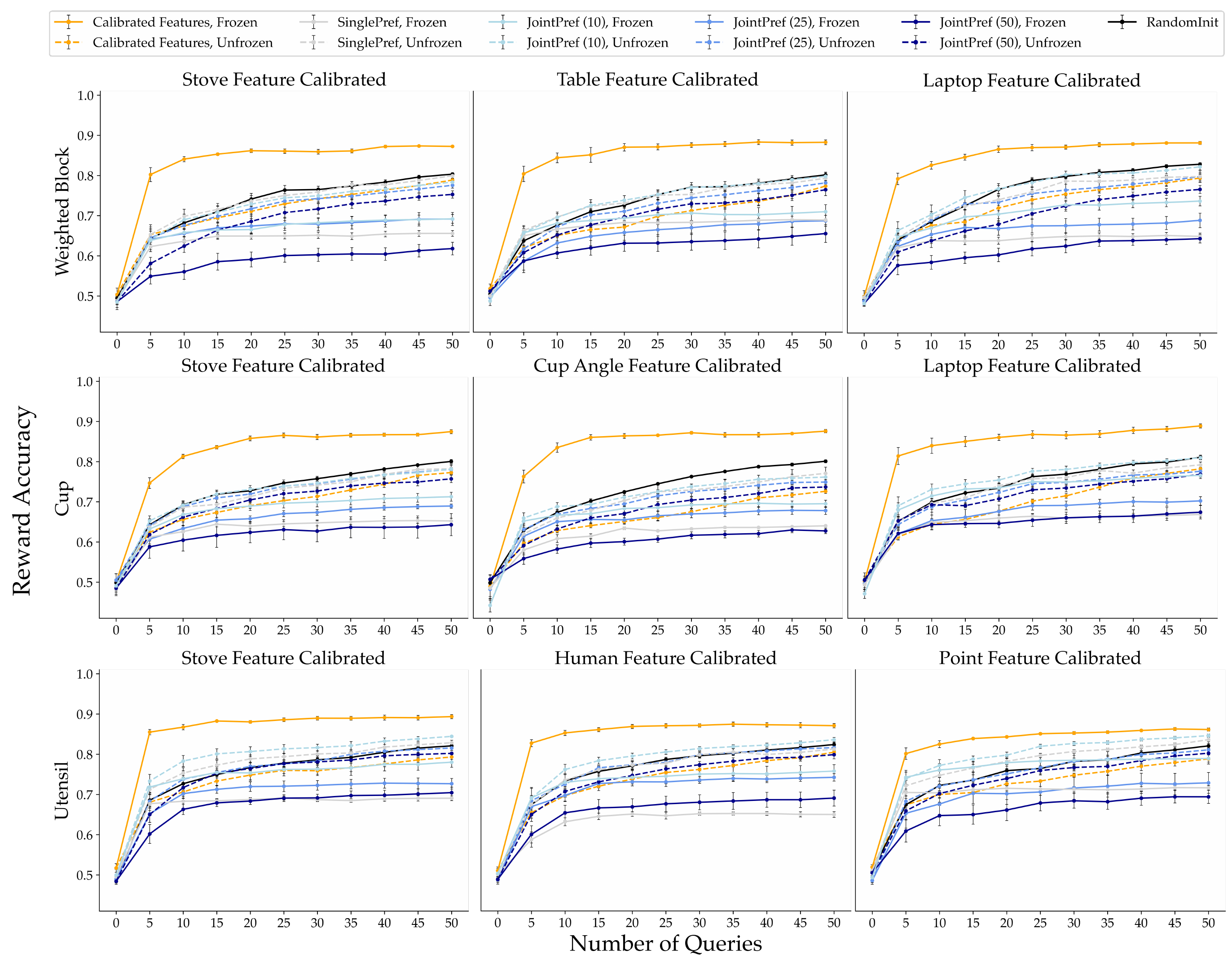}
    \caption{Single contextually-affected feature experiments shown with frozen and unfrozen representations.}
    \label{fig:rew-single-calib}
    \Description{Line plots showing reward accuracy against number of reward preference queries. For each method a there are two lines: one with frozen representations and one with unfrozen representations. Our method performs better with frozen representations while baselines generally perform better with unfrozen representations.}
    \vspace{15mm}
\end{figure*}

\begin{figure*}[!h]
    \centering
    \includegraphics[width=0.8\linewidth, alt={Line plots showing reward accuracy against number of reward preference queries. For each method a there are two lines: one with frozen representations and one with unfrozen representations. Our method performs better with frozen representations while baselines generally perform better with unfrozen representations.}]{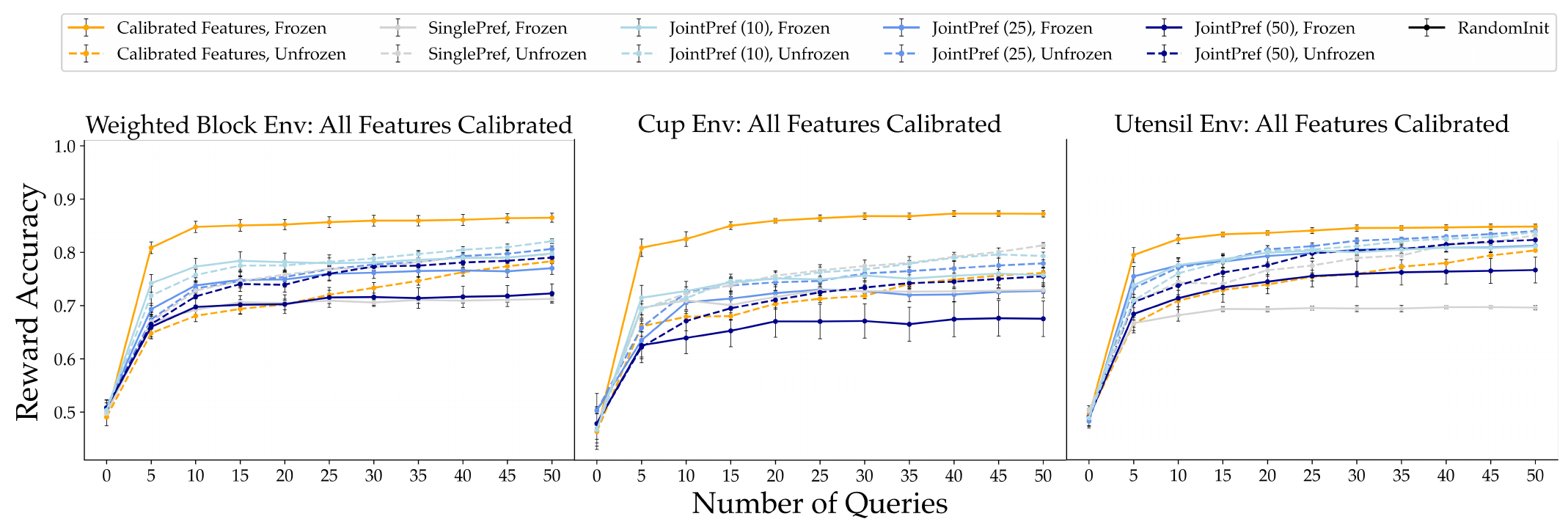}
    \caption{Multiple contextually-affected feature experiments shown with frozen and unfrozen representations.}
    \label{fig:rew-all-calib}
    \Description{Line plots showing reward accuracy against number of reward preference queries. For each method a there are two lines: one with frozen representations and one with unfrozen representations. Our method performs better with frozen representations while baselines generally perform better with unfrozen representations.}
\end{figure*}

We compare reward learning when pre-trained representations are either frozen or fine-tuned (Fig.~\ref{fig:rew-all-calib}\&~\ref{fig:rew-single-calib}). We observe that freezing the learned calibrated feature representations consistently yields higher reward accuracy, particularly in low-data regimes. This suggests that our method successfully learns meaningful intermediate features that can be reused in new rewards. In contrast, for the multi-task baseline, freezing and unfreezing perform similarly at low query counts, with frozen representations sometimes performing slightly better. However, as more data becomes available, fine-tuning the representation eventually leads to higher performance, indicating that the multi-task representation benefits from task-specific adaptation. In the main text we show the frozen variant of our method and the unfrozen variants of all multi-task baseline methods.

\section{User Study}
\subsection{Procedures} \label{app:user-procedures}
\begin{figure*}
    \centering
    \includegraphics[width=0.7\linewidth, alt={Screenshot of prompt for user description of personal contextual preference shown to participants.}]{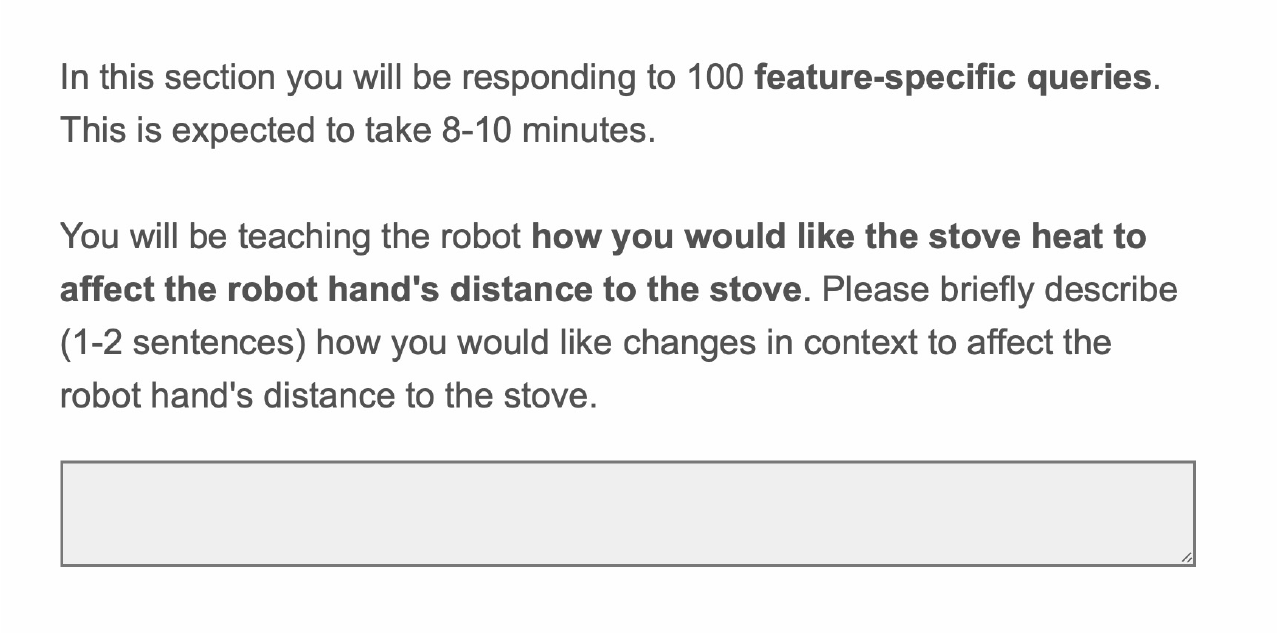}
    \caption{Example contextual preference prompt. Short answer descriptions of personal preferences were collected. Despite general priming, participants showed variety in responses.}
    \label{fig:user-prompt}
    \Description{Screenshot of prompt for user description of personal contextual preference shown to participants.}
\end{figure*}

\begin{figure*}
    \centering
    \includegraphics[width=0.7\linewidth, alt={Example contextual feature query language. Query stayed constant throughout round and was displayed next to query response GUI.}]{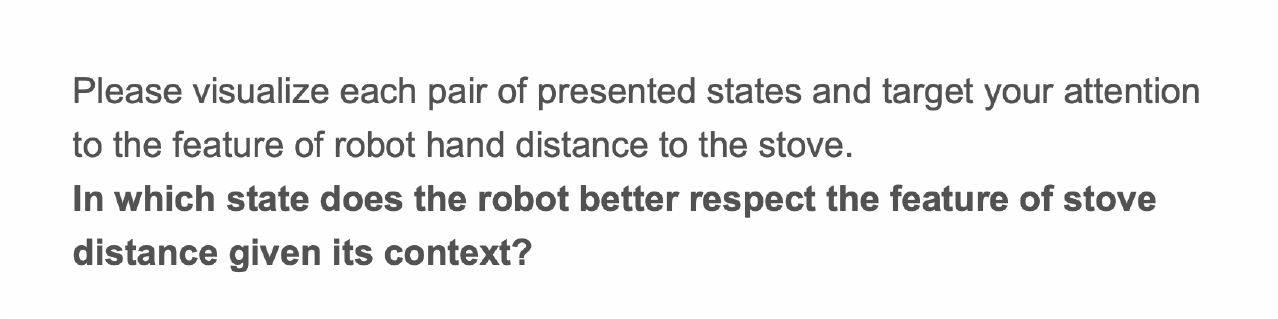}
    \caption{Example contextual feature query language. Query stayed constant throughout round and was displayed next to query response GUI.}
    \label{fig:user-query}
    \Description{Screenshot of contextual feature query language shown to participants.}
\end{figure*}

At the start of each query response round, participants were primed with a general preference, prompting them to teach the effects of a specific contextual element on a specific base feature. Participants were then asked to write down their personal contextual preference (see prompt example for \textit{stove} feature in Fig.~\ref{fig:user-prompt}). While our method supports accounting for the contextual effects of numerous contextual elements, we used this prompting to steer participants to teach the contextual effects of a single element such that we could visualize changes in the affecting contextual element when they visualized the learned models. The query response graphical user interface (GUI) was then pulled up (Fig.~\ref{fig:user-gui_query}), and participants were prompted with a contextual feature query (Fig.~\ref{fig:user-query}) which remained constant during the query round. 

The same state pairs were shown all participants for training both test features in the discretized \textit{Utensil} environment (see discretization details required for visualization in App.~\ref{sec:discrete-context}). Additionally, the laptop location was represented in the state, but when collecting queries, the irrelevant laptop was not visualized in order to prevent cluttering of the visualization. 

After completing a round of query response, participants completed post-query questionnaires including the SUS and NASA-TLX. Three calibrated feature models were trained while the participant took a 1-2 minute break. The participant then inspected the three learned models and the base feature (Fig.~\ref{fig:user-gui_inspect}), responded to the Likert scale for each model, and ranked the four models. 

\para{Feature order counterbalancing.} We accounted for the order of the two experimental features and the order in which trained model variants were presented to users. We counterbalanced the ordering of the two experimental features: half of the participants saw one feature first, and the other half saw the other feature first. When inspecting learned calibrated features, participants were presented with four trained model variants ("Relationships" 1-4) which represented calibrated features trained on 0 (base feature), 25, 50, or all 100 labeled queries (not necessarily in that order). Within each subject, we wanted to ensure the that the mapping between the relationship number and the model variant changed between the three features taught (one in familiarization phase and two in experimental phase). For the first feature trained in the familiarization phase, which had only three associated calibrated models (0, 25, 50), there were 3! = 6 possible orderings of those values. For the second and third experimental features, which had four calibrated models each (0, 25, 50, 100), there were 4! = 24 possible orderings of those values. Fully counterbalancing the order for all three features would therefore have required too many total permutations. To manage this, we used all 6 possible orderings of the first training feature, with each ordering appearing twice (once in each of the two experimental feature orderings). The orderings of the second and third training features were selected to be unique across participants within each ordering condition, but were reused across the two conditions. We aimed for approximate balance by rotating which training queries appeared in each position and ensuring that each participant saw a diverse set of training query orders.

\begin{figure*}[!ht]
    \centering
    \includegraphics[width=0.8\linewidth, alt={Examples of two different user contextual preferences for the stove feature, depicted by the user's written preference description the learned calibrated feature. Learned calibrated features are evaluated at all states in a point cloud and shown at different levels of stove heat.}]{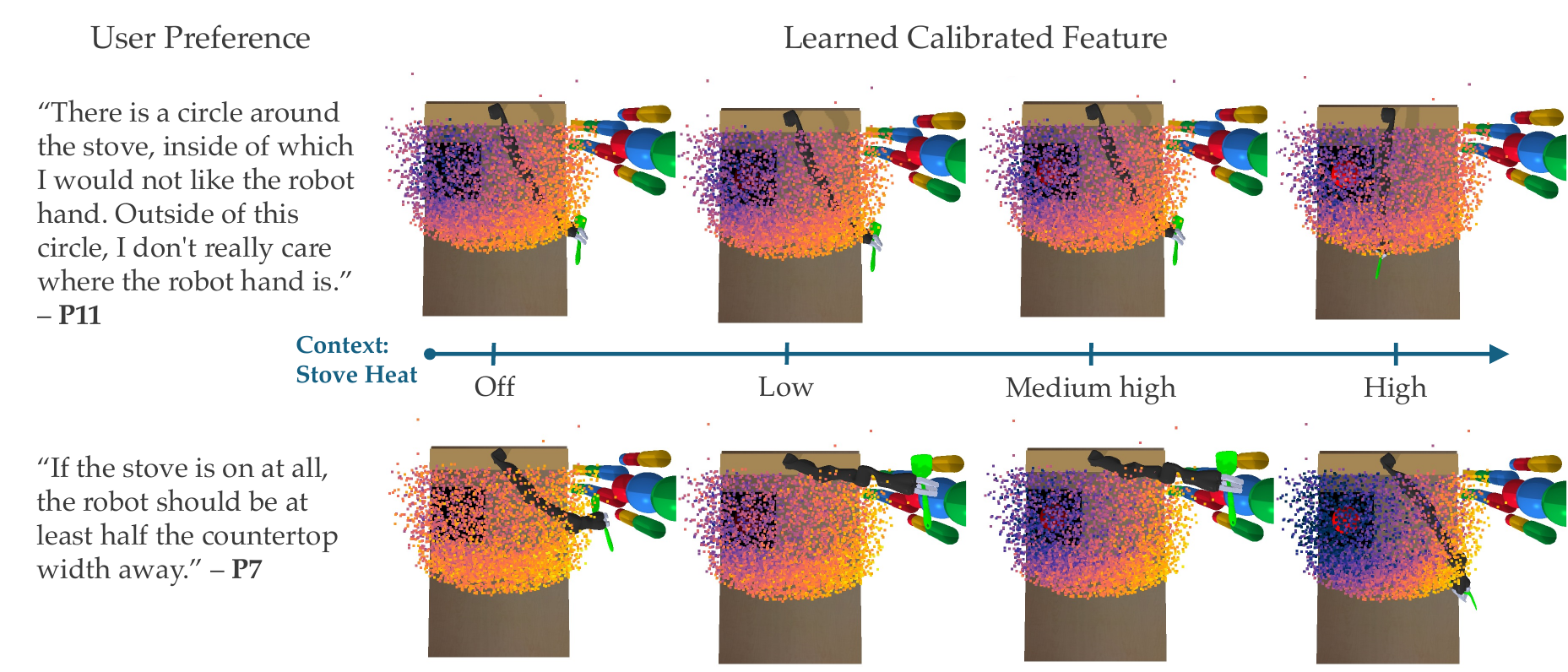}
    \caption{Differences in user preferences captured in learned calibrated features.}
    \label{fig:participant-learned-2}
    \Description{Examples of two different user contextual preferences for the stove feature, depicted by the user's written preference description the learned calibrated feature. Learned calibrated features are evaluated at all states in a point cloud and shown at different levels of stove heat.}
\end{figure*}

\subsection{Graphical user interfaces}  \label{app:user-guis}
\begin{figure*}
    \centering\includegraphics[angle=-90, width=0.65\linewidth, alt={Screenshot of graphical user interface. Shows manipulation environment robot and control panel.}]{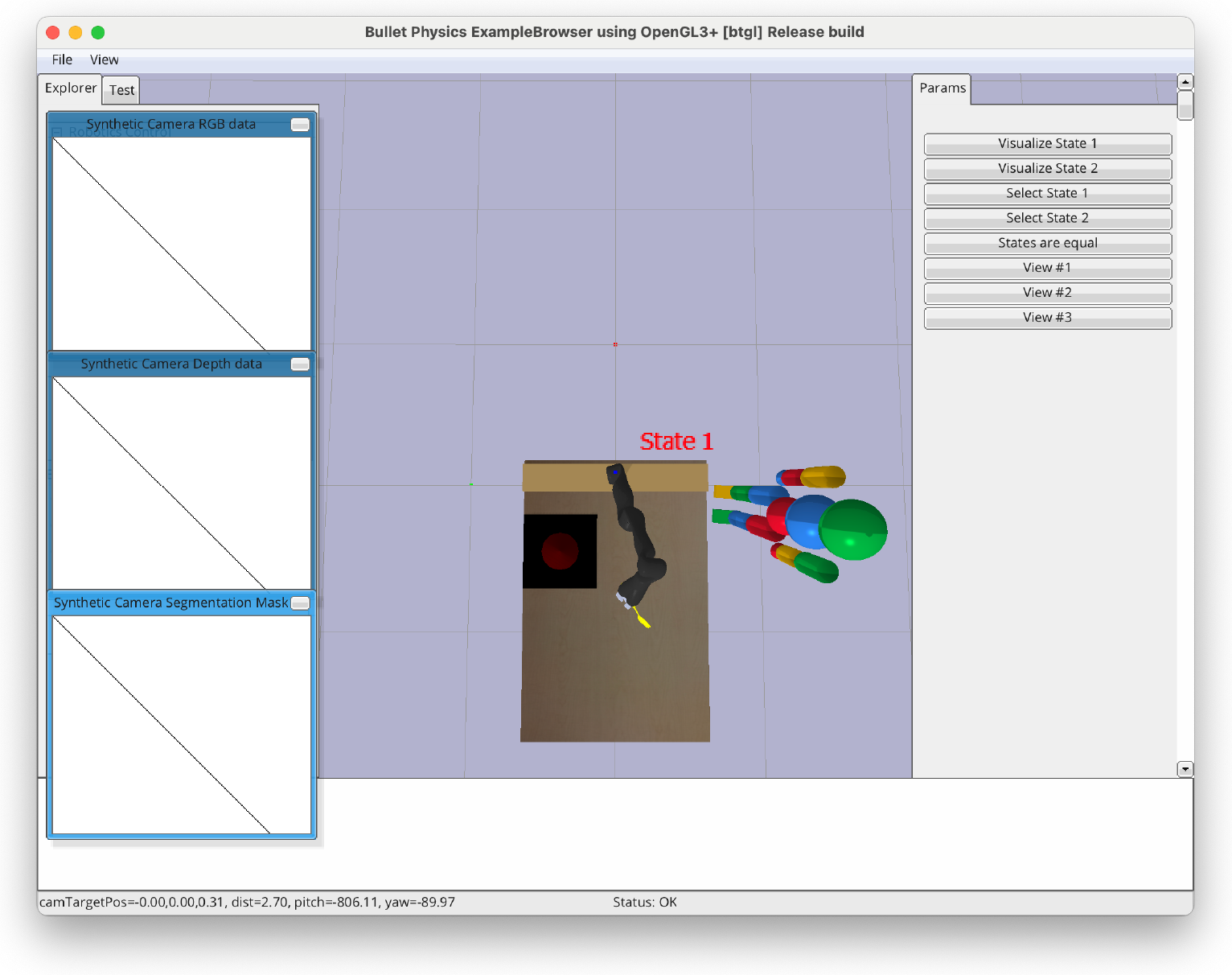}
    \caption{Query response GUI showing one state in pair. Participants used buttons on right panel to switch between visualizing two states and respond to queries. Left hand panel can be ignored.}
    \label{fig:user-gui_query}
    \Description{Screenshot of graphical user interface. Shows manipulation environment robot and control panel.}
\end{figure*}

\begin{figure*}
    \centering
    \includegraphics[angle=-90, width=0.65\linewidth, alt={Screenshot of graphical user interface. Shows point cloud of learned calibrated feature overlaid on manipulation environment and control panel.}]{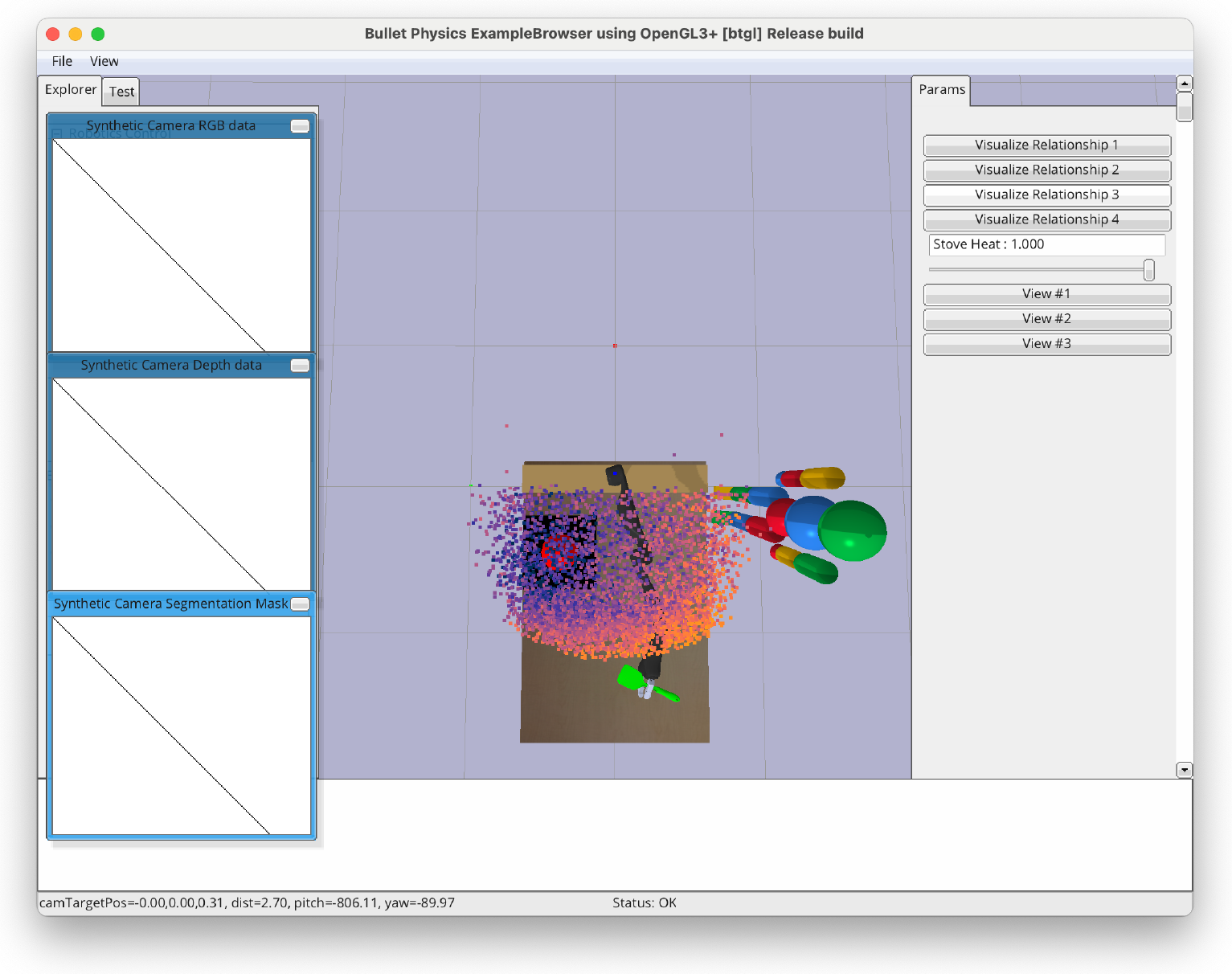}
    \caption{Model visualization GUI. Calibrated feature for \textit{stove} feature trained with 50 queries visualized in high heat context. Participants used buttons on right panel to select different models to visualize and sliding bar to move between four discrete contexts. Left hand panel can be ignored.}
    \label{fig:user-gui_inspect}
    \Description{Screenshot of graphical user interface. Shows point cloud of learned calibrated feature overlaid on manipulation environment and control panel.}
\end{figure*}

Participants interacted with two GUIs in order to both respond to contextual feature queries (Fig.~\ref{fig:user-gui_query}), and to interactively inspect learned calibrated features (Fig.~\ref{fig:user-gui_inspect}). When responding to queries, participants were able to view the states from different angles, and in response to the visualized contextual feature query select between "Select State 1", "Select State 2", or "States are equal". Users could also switch back and forth between the two compared states, and the current state was visualized in order to help users stay aware. Additionally, environment objects, like utensils, were colored differently to help make relevant context immediately salient and speed up query response. 

 Learned models were visualized as described in Fig.~\ref{fig:calib_feats}; however, 5k states were used as opposed to 8k for speed of evaluation and visualization. Participants could switch between the four models, and visualize a model in varying contexts by dragging a sliding bar which snapped the visualized context to the nearest of four spread out discrete contexts (and changed the model visualization accordingly). Participants were instructed on helpful ways to investigate and compare models, such as inspecting a single model across contexts, and switching between all models to see how they compare in a single context. Participants provided no input in this GUI, instead using it to respond to a survey collecting the Likert scale evaluation for each model and model rankings.

\subsection{Preference analysis} \label{app:pref-analysis}
Each participant provided a short free-response description of their contextual preference prior to responding to contextual feature queries (see prompt in Fig.~\ref{fig:user-prompt}). Therefore, we collected 12 short free-response answers regarding contextual preferences for each of the two test features. A single reviewer manually reviewed the responses and conservatively grouped them based on the level of specificity provided (Tables~\ref{tab:user-pref-human}\&~\ref{tab:user-pref-stove}). Responses describing contextual preferences in general, high-level terms without specific details were grouped together, while responses including concrete details were grouped separately unless the details matched exactly. This inductive grouping was intended to get a preliminary understanding of preference diversity. We found 6 users shared a general preference for the \textit{human} feature, with 2 users sharing a similar variant and 4 users holding unique variants. For the \textit{stove} feature, we found 7 users shared a general preference, and the remaining 5 users described individual variations.

In Fig.~\ref{fig:participant-learned-2} we give another example similar to Fig.~\ref{fig:participant-learned}, where we demonstrate how learned calibrated features differ for users with different preferences. The main difference in this example is the context in which the stove is off--while for one user the feature is equally respected in all states when the stove is off, the other user still desires the robot to stay away from directly above the stove. 

\begin{table*}[htbp]
\centering
\caption{Grouped preferences for \textit{human} feature.}
\renewcommand{\arraystretch}{1.6}
\begin{tabularx}{\textwidth}{@{}clX@{}}
\toprule
Group ID & Group Description & Free Response Preference \\
\midrule
1 & General preference (n=6) & "I would like to make sure that a sharper utensil is further from me as possible. A dull utensil is ok if it is closer." \\
1 & General preference (n=6) & "The sharper the utensil, the further it should be from the person." \\
1 & General preference (n=6) & "I would like sharper objects to be further away from the person and less sharp objects can be closer." \\
1 & General preference (n=6) & "The sharper the utensil the further away" \\
1 & General preference (n=6) & "I think I would like sharper utensils to be farther away from the human." \\
1 & General preference (n=6) & "Utensil sharpness should correlate with the preferred distance to the person." \\
2 & Unique preference        & "Keep the knife very far away from human, fork can be at half the arms length. Spatula and spoon could be close but not too close (a hands length away)" \\
3 & Unique preference        & "The spoon and spatula should be fine wherever. I'd like the fork and knife to be outside a range around the person, after which I don't really care where they are." \\
4 & Unique preference        & "There should be a general safety zone around the near vicinity of the person that no utensil should cross (personal space). The sharper the object, the larger that safety zone should be." \\
5 & Unique preference        & "Distance for the spoon and spatula doesn't matter. The knife needs to be kept further than the fork but past a certain distance it doesn't matter how far they are." \\
6 & Specific shared preference (n=2) & "generally, the sharper the utensil the further away from the person it should be. Spoons and spatulas are roughly equal in sharpness." \\
6 & Specific shared preference (n=2) & "Sharper objects should be further away from the human. The spoon and spatula are equally sharp" \\
\bottomrule
\end{tabularx}
\label{tab:user-pref-human}
\end{table*}

\begin{table*}[htbp]
\centering
\caption{Grouped preferences for \textit{stove} feature.}
\renewcommand{\arraystretch}{1.6}
\begin{tabularx}{\textwidth}{@{}clX@{}}
\toprule
Group ID & Group Description & Free Response Preference \\
\midrule
1 & General preference (n=7) & "the robot's hand should be father away from the stove as the heat increases." \\
1 & General preference (n=7) & "If the stove is hot, I would like the utensil to be further away. If the stove is not hot, the utensil can be closer." \\
1 & General preference (n=7) & "The hotter the stove is the further away the hand should be" \\
1 & General preference (n=7) & "The robot hand should be further away from a hotter stove" \\
1 & General preference (n=7) & "Increasing stove heat should correlate with the preferred distance from the end effector to the stove." \\
1 & General preference (n=7) & "If the stove is on, the utensil should not be directly over it; if the stove is off, this is OK. If the stove is hot, the utensil should not be in relative proximity (tapers off with temperature)." \\
1 & General preference (n=7)       & "I would want the robot hand to be further away from the stove the hotter the stove is." \\
2 & Unique preference        & "If the stove heat is at the highest keep the EE furthest away. If the stove is on in general It should be a hands length away. If it's off the EE can be anywhere" \\
3 & Unique preference        & "If the stove is on at all, the robot should be at least half the countertop width away." \\
4 & Unique preference        & "Disregarding what material the utensil is made out of, I would say that the utensil (context) does not really matter. There is a circle around the stove, inside of which I would not like the robot hand. Outside of this circle, I don't really care where the robot hand is." \\
5 & Unique preference        & "I don't want the robot to be close to the stove. Past a certain distance, I don't care where the robot is. Close is probably where I would expect the robot to feel the heat." \\
6 & Unique preference        & "I would like the robot's hand to be further away from the stove the hotter the stove is except when there's a spatula in hand." \\
\bottomrule
\end{tabularx}
\label{tab:user-pref-stove}
\end{table*}

\end{document}